\documentclass[10pt,twocolumn,letterpaper]{article}

\usepackage[pagenumbers]{cvpr} %

\usepackage[dvipsnames]{xcolor}

\usepackage{multirow}
\usepackage{colortbl}
\usepackage{wrapfig}
\usepackage{amssymb}%
\usepackage{pifont}%

\newcommand{\cmark}{\ding{51}}%
\newcommand{\xmark}{\ding{55}}%

\usepackage{multicol}
\usepackage{multirow}
\usepackage{amsmath}
\usepackage{amssymb}
\usepackage{subcaption}
\usepackage{graphicx}
\usepackage{xspace}
\usepackage{enumitem}
\usepackage{colortbl}
\usepackage{wrapfig}
\usepackage{algpseudocode}
\usepackage{algorithm}

\newcommand{\algorithmindent}{\hspace{\algorithmicindent}\hspace{\algorithmicindent}\hspace{\algorithmicindent}}

\DeclareMathOperator*{\argmax}{arg\,max}

\definecolor{cvprblue}{rgb}{0.21,0.49,0.74}
\usepackage[pagebackref,breaklinks,colorlinks,citecolor=cvprblue]{hyperref}
\usepackage[accsupp]{axessibility} %

\def\paperID{6396} %
\def\confName{CVPR}
\def\confYear{2024}

\title{Learning to Count without Annotations}

\author{%
  Lukas Knobel\\
  University of Amsterdam\\
  {\tt\small lukasknbl@gmail.com} \\
  \and
  Tengda Han\thanks{Equal senior contribution.}\\
  University of Oxford\\
  {\tt\small htd@robots.ox.ac.uk}
  \and
  Yuki M. Asano$^*$\\
  University of Amsterdam\\
  {\tt\small y.m.asano@uva.nl}
}
\newcommand{\midsepremove}{\aboverulesep = 0.2mm \belowrulesep = 0.5mm}
\midsepremove

\begin{document}
\maketitle
\begin{abstract}
While recent supervised methods for reference-based object counting continue to improve the performance on benchmark datasets, they have to rely on small datasets due to the cost associated with manually annotating dozens of objects in images. We propose UnCounTR, a model that can learn this task without requiring any manual annotations. To this end, we construct ``Self-Collages'', images with various pasted objects as training samples, that provide a rich learning signal covering arbitrary object types and counts. 
Our method builds on existing unsupervised representations and segmentation techniques to successfully demonstrate for the first time the ability of reference-based counting without manual supervision.
Our experiments show that our method not only outperforms simple baselines and generic models such as FasterRCNN and DETR, but also matches the performance of supervised counting models in some domains.\footnote{Code: \href{https://github.com/lukasknobel/SelfCollages}{https://github.com/lukasknobel/SelfCollages}}
\end{abstract}

\section{Introduction}

Cognitive neuroscientists  speculate that visual counting, especially for a small number of objects, is a pre-attentive and parallel process~\citep{trick1994small,dehaene2011number},
which can help humans and animals make prompt decisions~\citep{piazza2004number}.
Accumulating evidence shows that infants and certain species
of animals can differentiate between small numbers of
items~\citep{davis1988numerical,dehaene2011number,pahl2013numerical} and
as young as 18-month-old infants have been shown to develop counting abilities~\citep{slaughter2011learning}.
These findings indicate that the ability of visual counting may emerge very early or even be inborn in humans and animals.

On the non-biological side, recent developments in computer vision have been tremendous.
The state-of-the-art computer vision models can classify thousands of image classes~\citep{Krizhevsky2012alexnet,he2016deep}, detect various objects~\citep{zhou2022detecting}, or segment almost anything from an image~\citep{kirillov2023sam}.
Partially inspired by how babies learn to see the world~\citep{smith2005development}, 
some of the recent well-performing models are trained with self-supervised learning methods, whereby a learning signal for neural networks is constructed without the need for manual annotations~\citep{doersch2015unsupervised,he2020momentum}. %
The pretrained visual representations from such methods have demonstrated superior performances on various downstream visual tasks, like image classification and object detection~\citep{he2020momentum,caron2021emerging,he2022masked}. %
Moreover, self-supervised learning signals have been shown to be sufficient for successfully learning image groupings~\citep{yan2020clusterfit,vangansbeke2020scan} and even object and semantic segmentations without any annotations~\citep{caron2021emerging,zadaianchuk2023unsupervised}. %
Motivated by these, we ask in this paper whether visual counting might also be solvable without relying on human annotations.

\begin{figure}[t]
    \centering
    \includegraphics[width=0.23\textwidth]{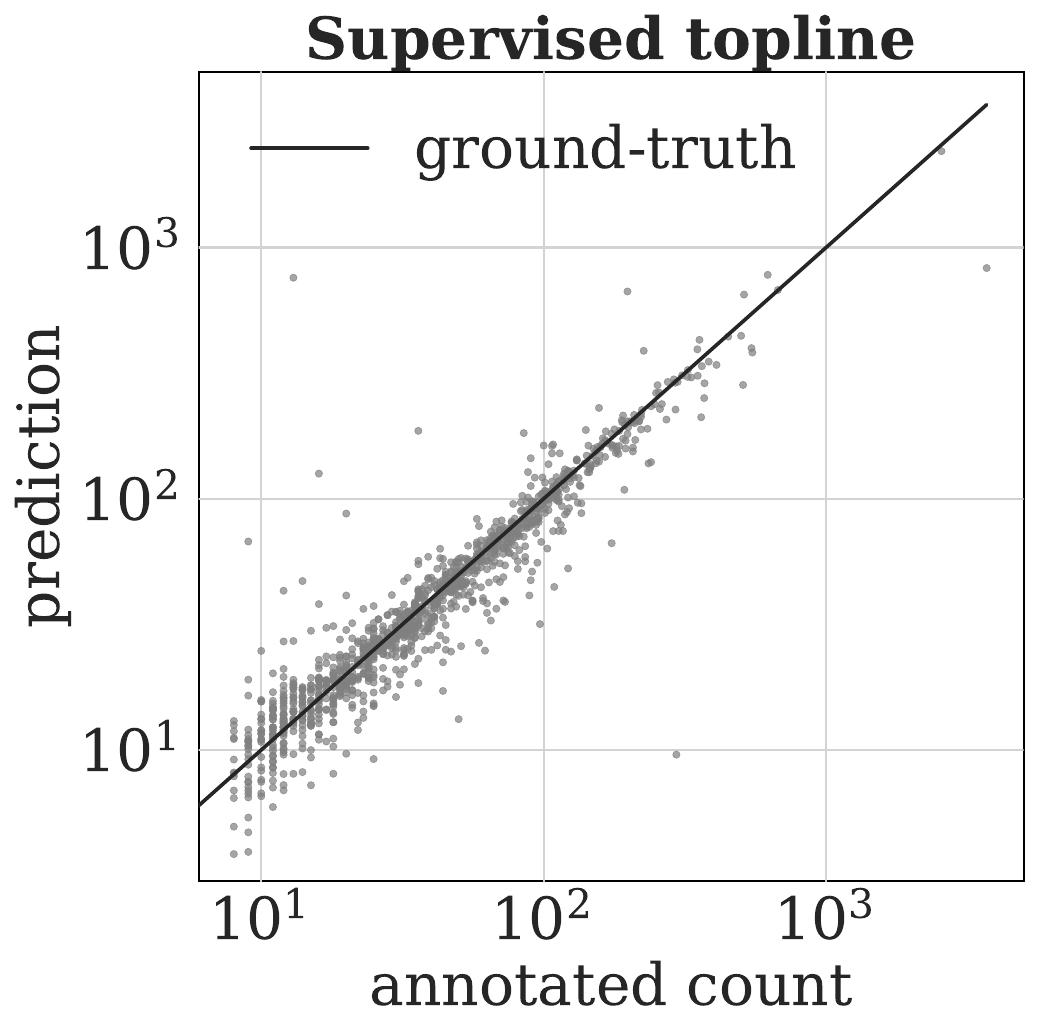}\includegraphics[width=0.23\textwidth]{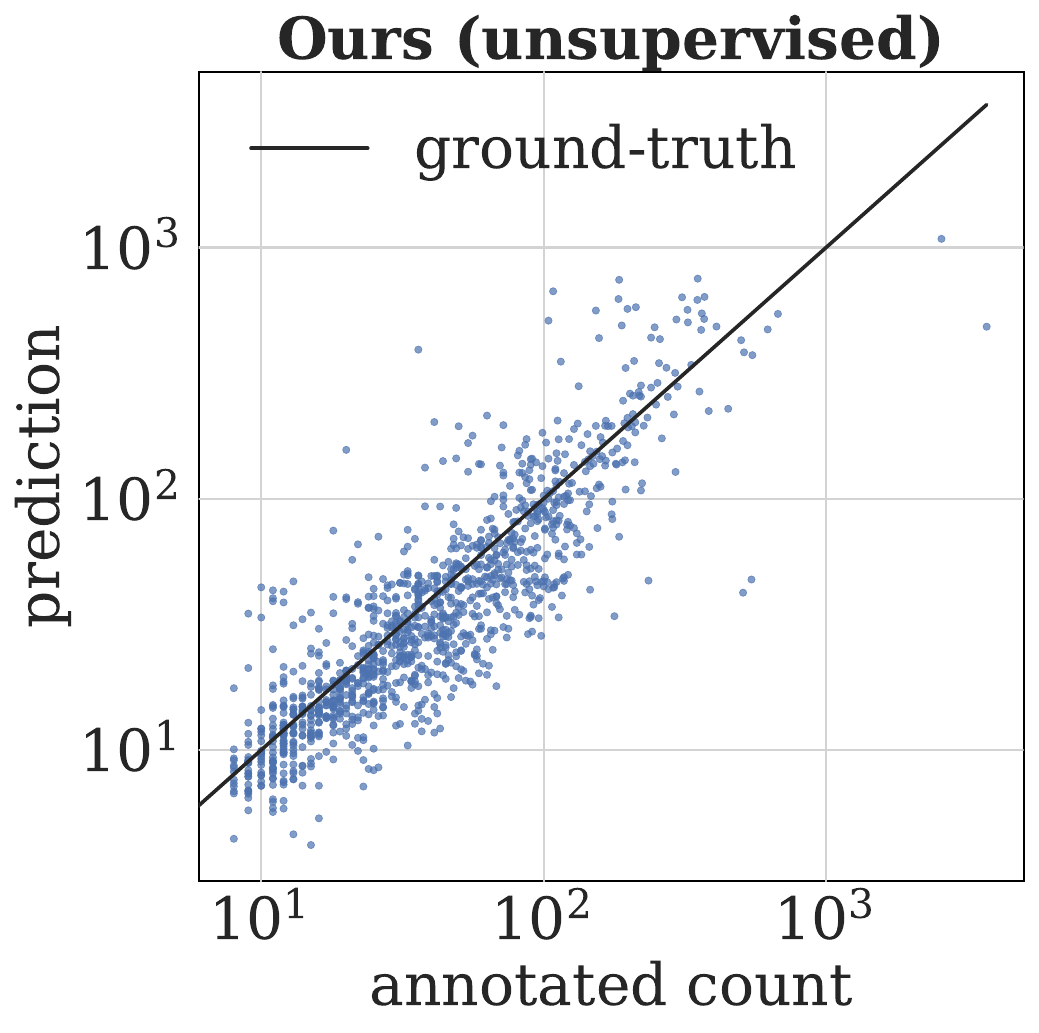}
    \caption{\textbf{CounTR \vs our proposed UnCounTRv2.} Our counting model, UnCounTRv2, is trained without any labels and manual counting annotations. It generalizes well from its up to 19 pseudo-counts encountered during training to the significantly higher numbers in the \mbox{FSC-147} test set. This demonstrates that learning to count is possible without annotations.}
    \label{fig:uncountrv2_countr}
    \vspace{-4mm}
\end{figure}
The current state-of-the-art visual counting methods,~\eg~CounTR~\citep{liu2022countr}, typically adapt pretrained visual representations to the counting task by using a considerable size of human annotations. %
However, we conjecture that the existing visual representations are already \emph{strong enough} to perform counting, even \emph{without} any manual annotations. 

In this paper, we design a straightforward self-supervised training scheme to teach the model `how to count', by pasting a number of objects on a background image, to make a Self-Collage.
Our experiments show that when constructing the Self-Collages carefully,
this training method is effective enough to leverage the pretrained visual representation on the counting task, even approaching other methods that require manually annotated counting data.
For the visual representation, we use the self-supervised pretrained DINO features~\citep{caron2021emerging}, which have been shown to be useful and generalisable for a variety of visual tasks like segmenting objects~\citep{melas2022deep,ziegler2022self}.
Note that the DINO model is also trained without manual annotations, 
thus our entire pipeline does not require annotated datasets.

\begin{figure*}
    \centering
    \includegraphics[width=\textwidth]{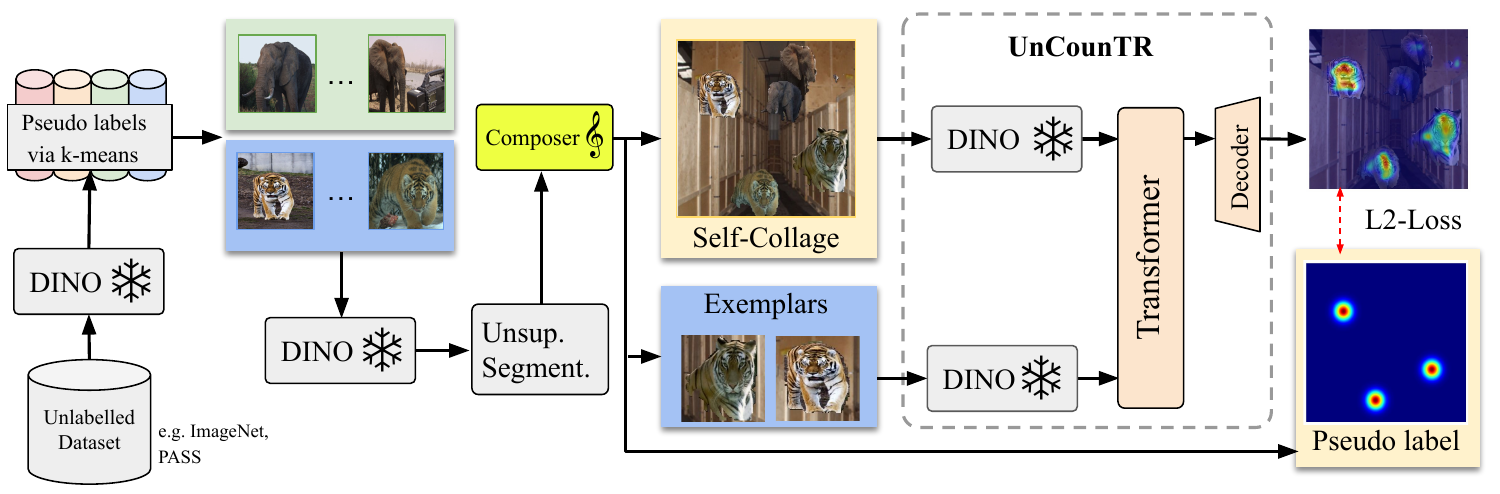}
    \vspace{-5mm}
    \caption{\textbf{Method overview.} Our method leverages the strong coherence of deep clusters to provide pseudo-labelled images which are used to construct a self-supervised counting task. The composer utilises self-supervised segmentations for pasting a set of objects onto a background image and UnCounTR is trained to count these when provided with unsupervised exemplars.
    }
    \vspace{-4mm}
    \label{fig:schematic}
\end{figure*}

To summarise, this paper focuses on the objective of~\emph{training a reference-based counting model without any manual annotation}.
The following contributions are made:
(i) We propose a simple yet effective data generation method to construct `Self-Collages', which pastes objects onto an image and gets supervision signals for free.
(ii) We leverage self-supervised pretrained visual features from DINO and develop UnCounTR, a transformer-based model architecture for counting.
(iii) The experiments show that our method trained without manual annotations not only outperforms baselines and generic models like FasterRCNN and DETR, but also matches the performance of supervised counting models.

\section{Related work}

\textbf{Counting with object classes.}
The class-specific counting methods are trained to count instances of a single class of interest, such as cars~\citep{mundhenk2016large, hsieh2017drone} or people~\citep{liu2018leveraging, sam2020completely, liang2023crowdclip, sam2019almost, liang2023crowdclip}. 
These methods require retraining to apply them to new object classes~\citep{mundhenk2016large}. In addition, some works rely on class-specific assumptions such as the proportionality of features and counts~\cite{katsuki2016unsupervised}, or the distribution~\citep{sam2020completely} or shapes~\citep{ubbens2020autocount} of objects, which cannot be easily adapted.

By contrast, class-agnostic approaches are not designed with a specific object class in mind. Early work by \citet{zhang2015salient} proposes the salient object subitizing task, where the model is trained to count by classification of $\{0,1,2,3,4+\}$ salient objects regardless of their classes.
Other reference-less methods like \citet{hobley2022learning} frame counting as a repetition-recognition task and aim to automatically identify the objects to be counted.
An alternative approach of class-agnostic counting requires a prior of the object type to be counted in the form of reference images, also called `exemplars', each containing a single instance of the desired class~\citep{yang2021class,lu2018class,djukic2022low,shi2023training, you2023few,ranjan2021learning, lin2021object}. %

\textbf{Counting with different methods.}
Categorised by the approach taken to obtain the final count,
counting methods can be divided into 
classification, detection and regression-based methods.

Classification-based approaches predict a discrete count for a given image~\citep{zhang2015salient}. The classes either correspond to single numbers or potentially open-ended intervals. Thus, predictions are limited to the pre-defined counts and a generalisation to new ranges is by design not possible.

An alternative is detection-based methods~\citep{hsieh2017drone}. By predicting bounding boxes for the counted instances and deriving a global count based on their number, these methods are unlike classification-based approaches not constrained to predefined count classes. While the bounding boxes can facilitate further analysis by explicitly showing the counted instances, the performance of detection-based approaches deteriorates in high-density settings~\citep{hobley2022learning}. 

Lastly, regression-based methods predict a single number and can be further divided into scalar and density-based approaches. Scalar methods directly map an input image to a single scalar count corresponding to the number of objects in the input~\citep{hobley2022learning}. Density-based methods on the contrary predict a density map for a given image and obtain the final count by integrating over it~\citep{lu2018class, liu2018leveraging, sam2020completely, djukic2022low, chen2022self, shi2019counting}. Similar to detection-based approaches, these methods allow locating the counted instances which correspond to the local maxima of the density map but have the added benefit of performing well in high-density applications with overlapping objects~\citep{lu2018class}. 
The recent work CounTR~\citep{liu2022countr} is a class-agnostic, density-based method which is trained to count both with and without exemplars using a transformer decoder structure with learnable special tokens.

\textbf{Self-supervised learning.}
Self-supervised learning (SSL) has shown its effectiveness in many computer vision tasks. Essentially, SSL derives the supervision signal from the data itself rather than manual annotations.
The supervision signal can be found from various ``proxy tasks'' like colorization~\citep{zhang2016colorful}, spatial ordering or impaining~\citep{doersch2015unsupervised,pathak2016context,he2022masked,noroozi2016jigsaw}, temporal ordering~\citep{misra2016shuffle,han2019video}, contrasting similar instances~\citep{oord2018representation,chen2020simple,he2020momentum}, clustering~\citep{caron2018deep,asano2019self}, and from multiple modalities~\citep{radford2021clip,alayrac2020mmv}.
Another line of SSL methods is knowledge distillation, where one smaller student model is trained to predict the output of the other larger teacher model~\citep{bucilua2006model,chen2017learning,kim2016sequence}. 
BYOL~\citep{grill2020bootstrap} design two identical models but train one model to predict the moving average of the other as the supervision signal.
Notably, DINO~\citep{caron2021emerging} is trained in a similar way but using Vision Transformers (ViTs) as the visual model~\citep{dosovitskiy2020image}, and obtains strong visual features. Simply thresholding the attention maps and using the resulting masks yields high-quality image segmentations. Follow-up works \citep{melas2022deep,shin2022unsupervised,ziegler2022self} demonstrate the semantic segmentation quality can be further improved by applying some light-weight training or post-processing of DINO features.
For counting in SSL, \citet{noroozi2017representation} use feature ``counting'' as a proxy task to learn representations. %
In this work, we focus on counting itself and explore approaches to teach the model to count without manual supervision.

\section{Method}

We tackle the task of counting objects given some exemplar crops within an image.
With an image dataset $\mathcal{D} = \{(\textit{\textbf{I}}_1, \mathcal{S}_1, \textbf{y}_1),\dots,(\textit{\textbf{I}}_i, \mathcal{S}_i, \textbf{y}_i)\}$, where $\textit{\textbf{I}}_i \in \mathbb{R}^{H\times W\times 3}$ denotes image $i$,
$\mathcal{S}_i=\{{\textit{\textbf{E}}_i}^{1},...,{\textit{\textbf{E}}_i}^{E}\}$ represent the $E$ visual exemplars of shape ${\textit{\textbf{E}}_i}^{e}\in\mathbb{R}^{H'\times W'\times 3}$, and $\textbf{y}_i\in\mathbb{R}^{H \times W}$ is the ground-truth density map,
the counting task can be written as:
\begin{equation}\label{eq:problem}
    \mathbf{\hat{y}}_i = f_{\Theta}(\textit{\textbf{I}}_i, \mathcal{S}_i)
\end{equation}
Here, the predicted density map $\mathbf{\hat{y}}_i$ indicates the objects to be counted, as specified by the exemplars $\mathcal{S}_i$, such that $\sum_{kl} \mathbf{\hat{y}}_{i_{kl}}$ yields the overall count for the image $\textit{\textbf{I}}_i$.
We are interested in training a neural network $f$ parameterised by $\Theta$ to learn how to count based on the exemplars $\mathcal{S}_i$.
For the supervised counting methods~\citep{liu2022countr,lu2018class},
the network parameters $\Theta$ can be trained with the `(\texttt{prediction}, \texttt{ground-truth})' pairs: $(\mathbf{\hat{y}}_i, \textbf{y}_i)$. 
However, for self-supervised counting, the learning signal $\textbf{y}_i$ is not obtained from manual annotations, but instead from the data itself.

In this section, we introduce two essential parts of our method:
we start by presenting our data generation method for counting in \Cref{method:data}, 
that is the construction of the tuple $( \textit{\textbf{I}}_i, \mathcal{S}_i, \textbf{y}_i) $; 
then, we explain the UnCounTR model, 
\ie $f_{\Theta}$ in \Cref{eq:problem}, in \Cref{sec:uncountr}.
An overview of our method is provided in \Cref{fig:schematic}.

\subsection{Constructing Self-Collages}\label{method:data}

A key component of self-supervised training is the construction of a supervision signal without any manual annotations. 
In this paper, we generate training samples by pasting different images on top of a background. 
This is a well-known technique~\citep{arandjelovic2019object,zhao2022x}, which 
next to its conceptual simplicity provides great flexibility regarding the number and type of pasted images.
While other works combine annotated training images to enrich the training set~\citep{hobley2022learning, liu2022countr}, we use this idea to construct the \textit{whole} training set including unsupervised proxy labels, yielding self-supervised collages, or Self-Collages for short. 
Their generation process is described by a composer module $g$, which forms a distribution ${g(\mathcal{O}, \mathcal{B}) = p(\boldsymbol{\mathit{\tilde I}},\mathcal{S},\textbf{y}\mid\mathcal{O},\mathcal{B})}$  of constructed images $\boldsymbol{\mathit{\tilde I}}\in \mathbb{R}^{H \times W \times 3}$ along with unsupervised exemplars $\mathcal{S}$ and labels $\textbf{y}$. The process is based on two sets of unlabelled images $\mathcal{O}$ and $\mathcal{B}$ for object and background images.
$\mathcal{S}=\{\textit{\textbf{E}}\}^E,\ \textit{\textbf{E}} \in\mathbb{R}^{H'\times W'\times 3}$ is a set of $E\in \mathbb{N}$ exemplars and  $\textbf{y}\in \mathbb{R}^{H\times W}$ corresponds to the density map of $\boldsymbol{\mathit{\tilde I}}$. 

The composer module $g$ first randomly selects the number of distinct object categories ${n_c\sim U[t_{\text{min}}, t_{\text{max}}]}$ the first of which is taken as the target cluster. $t_{\text{min}}$ and $t_{\text{max}}$ control the minimum and maximum number of categories in a Self-Collage. To reduce the risk of overfitting to construction artefacts, we always construct images with $n_{\text{max}}$ objects and change the associated number of target objects $\sum_{ij} \textbf{y}_{ij}$ solely by altering the number of objects in the target cluster. This way, the number of pasted objects and, therefore, the number of artefacts is independent of the target count. The number of target objects $n_0=n\sim U[n_{\text{min}}, n_{\text{max}}-n_c+1]$ has an upper-bound lower than $n_{\text{max}}$ to guarantee that there is at least one object of each of the $n_c$ types. For all other clusters, the number of objects is drawn from a uniform distribution of points on the $n_c-1$ dimensional polytope with $L1$-size of $n_{\text{max}}-n$ ensuring that the total number of objects equals $n_{\text{max}}$.  
Further details, including the pseudocode for the composer module, are shown in \Cref{sec:composer_details}.

\textbf{Unsupervised categories.} We obtain unsupervised object categories by first extracting feature representations for all samples in $\mathcal{O}$ using a pretrained DINO ViT-B/16 backbone~\citep{caron2021emerging} $d$ and subsequently running k-means with $K$ clusters:
\begin{equation}
    c(\textit{\textbf{I}}) = \text{k-means}_K{[(d(\mathcal{O}))^{\texttt{CLS}}]}(\textit{\textbf{I}}),
\end{equation}
where $c(\textit{\textbf{I}})$ is the unsupervised category for image $\textit{\textbf{I}}$ constructed using the final $\texttt{CLS}$-embedding of $d$.
For each of the $n_c$ clusters, a random, unique cluster $c_i,\ i\in[0, n_c-1]$ is chosen from all $K$ clusters where $c_0$ is the target cluster.

\textbf{Image selection.} Next, random sets of images $\mathcal{I}_i \subset \mathcal{O}$ are picked from their unsupervised categories $c_i$, such that  $|\mathcal{I}_i| = n_i$ and $c(\textit{\textbf{I}})=c_i\ \forall \textit{\textbf{I}} \in \mathcal{I}_i$. We denote the union of these sets as $\mathcal{I} = \bigcup_{i=0}^{n_c-1} \mathcal{I}_i$.
In addition, we sample one image $\textit{\textbf{I}}_\text{b}$ from another dataset $\mathcal{B}$, which is assumed to not contain salient objects to serve as the background image.

\subsubsection{Construction strategy} \label{sec:construction_strategy}
Here we detail the Self-Collage construction.
First, the background image  $\textit{\textbf{I}}_\text{b}$ is reshaped to the final dimensions $H\times W$ and used as a canvas on which the modified images $\mathcal{I}$ are pasted.
To mimic natural images that typically contain similarly sized objects, we first randomly pick a mean object size ${d_{\text{mean}}\sim U[d_\text{min}, d_\text{max}]}$. 
Subsequently, the target size of the pasted objects is drawn independently for each ${\textit{\textbf{I}}\in \mathcal{I}, \textit{\textbf{I}}\in \mathbb{R}^{d_{\text{h}} \times d_{\text{w}} \times 3}}$ from a uniform distribution $d_{\text{paste}}\sim U[(1-\sigma)\cdot d_\text{mean}, (1+\sigma)\cdot d_\text{mean}]$ where $\sigma\in (0,1)$ controls the diversity of objects sizes in an image. 
After resizing $\textit{\textbf{I}}$ to $\textit{\textbf{I}}_\text{r}\in \mathbb{R}^{d_{\text{paste}} \times d_{\text{paste}} \times 3}$, the image is pasted to a random location on $\textit{\textbf{I}}_\text{b}$. 
This location is either a position where previously no object has been pasted, or any location in the constructed image, potentially leading to overlapping images. By default, we will use the latter.

\textbf{Segmented pasting.} Since pasting the whole image $\textit{\textbf{I}}_\text{r}$ might violate the assumption of having a single object and results in artefacts by also pasting the background of $\textit{\textbf{I}}_\text{r}$, we introduce an alternative using self-supervised segmentations. 
This method, which we will use by default, uses an unsupervised segmentation method~\citep{shin2022unsupervised} to obtain a noisy foreground segmentation $\textbf{s}\in [0,1]^{d_{\text{h}}\times d_{\text{w}}}$ for $\textit{\textbf{I}}$. Instead of pasting the whole image, the segmentation $\textbf{s}$ is used to only copy its foreground. 
Additionally, having access to $\textbf{s}$, this method can directly control the size of the pasted objects rather than the pasted images. To do that, we first extract the object in $\textit{\textbf{I}}$ by computing the Hadamard product $\textit{\textbf{I}}_\text{object} = \text{cut}\left(\textit{\textbf{I}} \circ \textbf{s}\right)$ where the operation ``cut'' removes rows and columns that are completely masked out. 
In the next step, $\textit{\textbf{I}}_\text{object}\in \mathbb{R}^{h_\text{object}\times w_\text{object} \times 3}$ is resized so that its maximum dimension equals $d_\text{paste}$ before pasting it onto $\textit{\textbf{I}}_\text{b}$.

\textbf{Exemplar selection.} 
To construct the exemplars used for training, we exploit the information about how the sample was constructed using $g$:  The set of $E$ exemplars $\mathcal{S}$ is simply obtained by filtering for pasted objects that belong to the target cluster $c_0$ and subsequently sampling $E$ randomly. Then, for each of them, a crop of  $\boldsymbol{\mathit{\tilde I}}$ is taken as exemplar after resizing its spatial dimensions to $H'\times W'$.

\subsubsection{Density map construction} \label{sec:density_map_construction}
To train our counting model, we construct an unsupervised density map $\textbf{y}$ for each training image $\boldsymbol{\mathit{\tilde I}}$. 
It needs to have the following two properties: i) it must sum up to the overall count of objects that we are counting and ii) it must have high values at object locations.
To this end, we create  $\textbf{y}$ as a simple density map of Gaussian blobs as done in supervised density-based counting methods~\citep{djukic2022low, liu2022countr}. 
For this, we use the bounding box for each pasted target image $\textit{\textbf{I}}\in \mathcal{I}_0$ and place Gaussian density at the centre and normalise it to one.

\subsection{UnCounTR\label{sec:uncountr}}

\textbf{Model architecture.}
UnCounTR's architecture is inspired by CounTR~\citep{liu2022countr}.
To map an input image $\textit{\textbf{I}}$ and its exemplars $\mathcal{S}$ to a density map $\mathbf{\hat{y}}$, the model consists of four modules: image encoder $\Phi$, exemplar encoder $\Psi$, feature interaction module $f_{\text{fim}}$, and decoder $f_{\text{dec}}$. An overview of this architecture can be seen in \Cref{fig:schematic}.
The image encoder $\Phi$ encodes an image $\textit{\textbf{I}}$ into a feature map $\textbf{x}=\Phi(\textit{\textbf{I}})\in \mathbb{R}^{h \times w \times d}$ where $h,w,d$ denote the height, width, and channel depth.
Similarly, each exemplar $\textit{\textbf{E}} \in \mathcal{S}$ is projected to a single feature vector $\textbf{z} \in \mathbb{R}^{d}$ by taking a weighted average of the grid features. %
Instead of training a CNN for an exemplar encoder as CounTR does, 
we choose $\Psi=\Phi$ to be the frozen DINO visual encoder weights.
Reusing these weights, the exemplar and image features are in the same feature space.
The feature interaction module (FIM) $f_{\text{fim}}$ enriches the feature map $\textbf{x}$ with information from the encoded exemplars $\textbf{z}_j,\ j \in \{1,...,E\}$ with a transformer decoder structure.
Finally, the decoder $f_{\text{dec}}$ takes the resulting patch-level feature map of the FIM as input and upsamples it with 4 convolutional blocks, ending up with a density map of the same resolution as the input image.
Please refer to \Cref{sec:model_architecture_appendix} for the full architectural details.

\textbf{UnCounTR supervision.}
UnCounTR is trained using the Mean Squared Error between model prediction and pseudo labels. Given a Self-Collage $\boldsymbol{\mathit{\tilde I}}$, exemplars $\mathcal{S}$, and density map $\textbf{y}$, the loss $\mathcal{L}$ for an individual sample is computed using the following equation where $f_{\Theta}(\boldsymbol{\mathit{\tilde I}}, \mathcal{S})_{ij}$ is UnCounTR's spatially-dense output at location $(i,j)$:
\begin{equation}
    \mathcal{L} = \frac{1}{H\cdot W} \sum_{ij} (\textbf{y}_{ij} - f_{\Theta}(\boldsymbol{\mathit{\tilde I}}, \mathcal{S})_{ij})^2
\end{equation}

\section{Experiments}

\subsection{Implementation Details}\label{sec:implementation_details}
\textbf{Datasets. }
To construct Self-Collages, we use \textbf{ImageNet-1k}~\citep{deng2009imagenet} and \textbf{SUN397}~\citep{xiao2010sun}.
ImageNet-1k contains 1.2M mostly object-centric images spanning 1000 object categories.
SUN397 contains 109K images for 397 scene categories like `cliff' or `corridor'. 
Note that the object or scene category information is never used in our method.
We assume that images from ImageNet-1k contain a single salient object and images from SUN397 contain no salient objects to serve as sets $\mathcal{O}$ and $\mathcal{B}$. 
Based on this, $g$ randomly selects images from SUN397 as background images and picks objects from ImageNet-1k.
Although Imagenet-1k and SUN397 contain on average 3 and 17 objects respectively~\citep{lin2014microsoft}, these assumptions are still reasonable for our data construction.
Additional ablations show that even explicitly violating the assumption of having no salient objects in $\mathcal{B}$ by using $\mathcal{O}=\mathcal{B}$ only have a small effect on performance, indicating the robustness of our method against violations (see \Cref{sec:dataset_ablations}).
An example of a Self-Collage with 5 objects can be seen in~\Cref{fig:schematic}.
Examples of both datasets and a more detailed discussion of these assumptions are included in \Cref{sec:self_collage_details}.

\begin{table*}[tb]
    \centering
    \small
    \vspace{-2mm}
    \setlength{\tabcolsep}{0.4em}
    \begin{tabular}{l @{\hskip 0.5em} ccc c ccc c ccc }
        \toprule
        \multirow{2}*{Method}  & \multicolumn{3}{c}{FSC-147 \textit{low}}  && \multicolumn{3}{c}{FSC-147 \textit{medium}} && \multicolumn{3}{c}{FSC-147 \textit{high}} \\
        \cmidrule{2-4} \cmidrule{6-8} \cmidrule{10-12} 
        & MAE$\downarrow$ & RMSE$\downarrow$ & $\tau\uparrow$ && MAE$\downarrow$ & RMSE$\downarrow$ & $\tau\uparrow$ &&  MAE$\downarrow$ & RMSE$\downarrow$ & $\tau\uparrow$\\
        \midrule
        Average & {37.71} & {37.79} & - && {22.74} & {23.69} & - && {68.88} & {213.08} & - \\
        Conn. Comp. &  {14.71} &  {18.59} &  {0.14} &&  {14.19} &  {17.90} &  {0.16} &&  {69.77} &  {210.54} &  {0.17}\\
        FasterRCNN & 7.06 & {8.46} & -0.03 && 19.87 & 22.25 & -0.04 && 109.12 & 230.35 & -0.06\\
        DETR & 6.92 & \textbf{8.20} & 0.07 && 19.33 & 21.56 & -0.08 && 109.34 & \textbf{162.88} & -0.07\\
        \midrule
        \textbf{UnCounTR (ours)}  & \textbf{5.60} & 10.13 & \textbf{0.27} && \textbf{9.48} & \textbf{12.73} & \textbf{0.34} && \textbf{67.17} & {189.76}  & \textbf{0.26}\\
        \multicolumn{1}{r}{$\sigma\text{(5 runs)}$}&\,${\pm0.48}$ & $\,{\pm0.84}$ & $\,{\pm0.02}$ && $\,{\pm0.19}$ & $\,{\pm0.33}$ & $\,{\pm0.02}$ && $\,{\pm1.03}$ & $\,{\pm1.38}$  & $\,{\pm0.01}$\\
        \bottomrule
    \end{tabular}
    \vspace{-2mm}
    \caption{\textbf{Comparison to baselines.} Evaluation on different FSC-147 test subsets. ``Conn. Comp.'' refers to connected components.}
    \label{tabl:FSC147_baselines}
    \vspace{-2mm}
\end{table*}

To evaluate the counting ability, we use the standard \textbf{FSC-147} dataset~\citep{ranjan2021learning},
which contains 6135 images covering 147 object categories,
with counts ranging from 7 to 3731 and an average count of 56 objects per image.
For each image, the dataset provides at least three randomly chosen object instances with annotated bounding boxes as exemplars. 
To analyse the counting ability in detailed count ranges,
we partition the FSC-147 test set into three parts of roughly 400 images each, resulting in \textit{FSC-147-\{low,medium,high\}} subsets, each covering object counts from 8-16, 17-40 and 41-3701 respectively.
Unless otherwise stated, we evaluate using 3 exemplars.

Additionally, we also use the \textbf{CARPK}~\citep{hsieh2017drone} and the Multi-Salient-Object (\textbf{MSO}) dataset~\citep{zhang2015salient} for evaluation. CARPK consists of 459 images of parking lots. Each image contains between 2 and 188 cars, with an average of 103.
MSO contains 1224 images covering 5 categories: \{0,1,2,3,4+\} salient objects with bounding box annotations.
This dataset is largely imbalanced as 338 images contain zero salient objects and 20 images contain at least 4.
For evaluation, we removed all samples with 0 counts, split the 4+ category into exact counts, and chose only one annotated object as exemplar.

\textbf{Construction details. }
We configure the composer module to construct training samples using objects of $K=10,000$ different clusters. To always have objects of a target and a non-target cluster in each image, we set $t_\text{min}=t_\text{max}=2$. Since the minimum number of target objects in an image limits the maximum number of exemplars available during training, we set $n_\text{min}=3$, the maximum is $n_\text{max}=20$. Finally, we choose $d_\text{min}=15$, $d_\text{max}=70$, and $\sigma=0.3$ to obtain diverse training images with objects of different sizes as confirmed by visual inspection of the generated Self-Collages (see \Cref{sec:self_collage_examples}).

\textbf{Training and inference details. }
During training, the images are cropped and resized to $224\times224$ pixels. The exemplars are resized to $64\times64$ pixels. For each batch, we randomly draw the number of exemplars $E$ to be between 1 and 3. We follow previous work~\citep{liu2022countr} by scaling the loss, in our case by a factor of 3,000, and randomly dropping 20\% of the non-object pixels in the density map to decrease the imbalance of object and non-object pixels. By default, we use an AdamW optimizer and a cosine-decay learning rate schedule with linear warmup and a maximum learning rate of $5\times10^{-4}$. We use a batch size of 128 images. Each model is trained on an Nvidia A100 GPU for $50$ epochs of $10,000$ Self-Collages each, which takes about $4$ hours. At inference, we use a sliding window approach similar to \citet{liu2022countr}, see \Cref{sec:inference_details} for details.
For evaluation metrics, we report Mean Absolute Error (MAE) and Root Mean Squared Error (RMSE) following previous work~\citep{liu2022countr, djukic2022low}. We also report Kendall's $\tau$ coefficient, which is a rank correlation coefficient between the sorted ground-truth counts and the predicted counts.

\textbf{Baselines. }
To verify the effectiveness of our method, we introduce a series of baselines to compare with.
(1) \textbf{Average} baseline: the prediction is always the average count of the FSC-147 training set, which is 49.96 objects.
(2) \textbf{Connected components} baseline: for this, we use the final-layer attention values from the \texttt{CLS}-token of a pretrained DINO ViT-B/8 model. To derive a final count, we first threshold the attention map of each head to keep $p_\text{att}$ percent of the attention mask. Subsequently, we consider each patch to where at least $n_\text{head}$ attention heads attend to belong to an object. The number of connected components in the resulting feature map which cover more than $p_\text{size}$ percent of the feature map is taken as prediction. To this end, we perform a grid search with almost 800 configurations of different value combinations for the three thresholds $p_\text{att}$, $n_\text{head}$, and $p_\text{size}$ on the FSC-147 training set and select the best configuration based on the MAE. The specific values tested can be found in \Cref{sec:connected_components_appendix}.
(3) \textbf{FasterRCNN} baseline:
we run the strong image detection toolbox FasterRCNN~\citep{ren2015faster} with a score threshold of 0.5 on the image, which predicts a number of object bounding boxes. Then the object count is obtained by parsing the detection results by taking the total number of detected bounding boxes. 
Just like the connected components baseline, this model is applied to images resized to $384\times384$ pixels. 
(4) \textbf{DETR} baseline:
similar to FasterRCNN, we evaluate the detection model DETR~\citep{carion2020end} on the counting task. Here, we resize the images to $800\times800$ pixels to better match DETR's evaluation protocol.

\begin{table*}
\setlength{\tabcolsep}{2.4pt}
\vspace{-1mm}
\begin{subtable}[h]{0.45\textwidth}
    \centering
    \small
    \begin{tabular}{ c c @{\hskip 0.5em} ccc cc }
        \toprule
         &   & \multicolumn{2}{c}{FSC-147}  && \multicolumn{2}{c}{FSC-147 \textit{low}}\\
        \cmidrule{3-4} \cmidrule{6-7}
        Frozen $\Phi$ & Frozen $\Psi$ & MAE$\downarrow$ & RMSE$\downarrow$ && MAE$\downarrow$ & RMSE$\downarrow$\\
        \midrule
        \xmark & \xmark & 37.64 & \textbf{126.91} && 7.99 & 13.37 \\
        \cmark & \xmark & 36.94 & 130.22 && 9.10 & 14.69 \\
        \rowcolor{lightgray} \cmark & \cmark &  \textbf{35.77} & 130.34  && \textbf{5.60} & \textbf{10.13}\\
        \bottomrule
    \end{tabular}
    \caption{Keeping both image encoder $\Phi$ and exemplar encoder $\Psi$ frozen works best.}
    \label{tabl:FSC147_ablation_model_arch}

\end{subtable}
\hfill
\begin{subtable}[h]{0.48\textwidth}

    \centering
    \small
    \begin{tabular}{ c c @{\hskip 0.5em} cc ccc }
        \toprule
         &    & \multicolumn{2}{c}{FSC-147}  && \multicolumn{2}{c}{FSC-147 \textit{low}}\\
        \cmidrule{3-4} \cmidrule{6-7}
        Segm. & Overlap. & MAE$\downarrow$ & RMSE$\downarrow$ && MAE$\downarrow$ & RMSE$\downarrow$\\
        \midrule
        \cmark & \xmark & 36.47 & \textbf{128.97} && 8.84 & 14.54 \\
        \xmark & \cmark  & 37.46 & 136.03 && 6.33 & 11.55 \\
        \rowcolor{lightgray} \cmark & \cmark & \textbf{35.77} & 130.34  && \textbf{5.60} & \textbf{10.13} \\
        \bottomrule
    \end{tabular}
    \caption{Using both segmented objects and allowing for overlapping objects works well.}
    \label{tabl:FSC147_ablation_construction}

\end{subtable}\\
\begin{subtable}[h]{0.45\textwidth}
\centering
\small
\begin{tabular}{ c @{\hskip 0.5em} cc c cc }
    \toprule
    & \multicolumn{2}{c}{FSC-147}  && \multicolumn{2}{c}{FSC-147 \textit{low}}\\
    \cmidrule{2-3} \cmidrule{5-6}
    $n_{\text{max}}$ & MAE$\downarrow$ & RMSE$\downarrow$ && MAE$\downarrow$ & RMSE$\downarrow$\\
    \midrule
    50 & \textbf{30.17} & \textbf{115.42} && 6.65 & 13.60  \\
    \rowcolor{lightgray} 20 &  35.77 & 130.34  && \textbf{5.60} & \textbf{10.13}\\
    \bottomrule
\end{tabular}
\caption{The number of pasted objects correlates with performance.}
\label{tabl:FSC147_ablation_num_objects}

\end{subtable}
\hfill
\begin{subtable}[h]{0.48\textwidth}
    \centering
    \small
    \begin{tabular}{ c @{\hskip 0.5em} cc c cc }
        \toprule
        & \multicolumn{2}{c}{FSC-147}  && \multicolumn{2}{c}{FSC-147 \textit{low}}\\
        \cmidrule{2-3} \cmidrule{5-6}
        \#Exemplars & MAE$\downarrow$ & RMSE$\downarrow$ && MAE$\downarrow$ & RMSE$\downarrow$\\
        \midrule
        0 - 3 & 36.28 & 131.44 && 6.23 & 10.33  \\
        \rowcolor{lightgray} 1 - 3 &  \textbf{35.77} & \textbf{130.34}  && \textbf{5.60} & \textbf{10.13}\\
        \bottomrule
    \end{tabular}
    \caption{Training our method strictly in the 1-3 exemplar setting works best.}
    \label{tabl:FSC147_ablation_num_exemplars}

\end{subtable}\\
\vspace{-2mm}
\caption{\textbf{Ablations.} We ablate various components of our model architecture and Self-Collage construction method. Default settings are highlighted in \colorbox{lightgray}{grey}.}
\vspace{-3mm}
\end{table*}

\begin{table}[tb]
    \vspace{-2mm}
    \centering
    \small
    \setlength{\tabcolsep}{0.2em}
    \begin{tabular}{ l l @{\hskip 0.5em} cc c cc }
        \toprule
         &   & \multicolumn{2}{c}{FSC-147}  && \multicolumn{2}{c}{FSC-147 \textit{low}}\\
        \cmidrule{3-4} \cmidrule{6-7}
        Architecture & Pretraining & MAE$\downarrow$ & RMSE$\downarrow$ && MAE$\downarrow$ & RMSE$\downarrow$\\
        \hline
        ViT-B/8 & Leopart \citep{ziegler2022self} & 36.04 & {126.75} && 5.72 & 10.67\\
        ViT-B/8 & DINO \citep{caron2021emerging} & 38.16 & 132.09 && 6.10 & {9.79}\\
        ViT-B/16 & DINO \citep{caron2021emerging} &  {35.77} & 130.34  && {5.60} & 10.13\\
        \bottomrule
    \end{tabular}
    \vspace{-2mm}
    \caption{\textbf{Combining Self-Collages with different backbones.} We can apply our method to any recent state-of-the-art pretrained network and achieve strong performances.}
    \label{tabl:FSC147_ablation_backbone}
    \vspace{-5mm}
\end{table}

\subsection{Comparison against baselines}
In~\Cref{tabl:FSC147_baselines}, we compare against multiple baselines on the three splits of FSC-147.
As the connected components cannot leverage the additional information of the few-shot samples provided, we try to make the comparison as fair as possible, by testing almost $800$ threshold parameters on the FSC-147 training set. 
While this yields a strong baseline, we find that our method of learning with Self-Collages more than halves the MAE on the \textit{low} split, despite using a similar visual backbone.
Next, we compare against the FasterRCNN and DETR object detectors which, unlike UnCounTR, are trained in a supervised fashion. Even though DETR outperforms UnCounTR in terms of RMSE on FSC-147 \textit{low} and \textit{high}, we find that our method still outperforms all baselines on 7 out of 9 measures. This is despite the additional advantages such as access to the FSC-147 training distribution for these baselines. 
The gap between most baselines and UnCounTR is the smallest on the \textit{high} split, suggesting limits to its generalisation which we will explore further in \Cref{sec:generatlisation_ood}.
There, we also provide a qualitative analysis of some failure cases of FasterRCNN.

\subsection{Ablation Study} \label{sec:ablation_study}
We analyse the various components of our method in~\Cref{tabl:FSC147_ablation_model_arch,tabl:FSC147_ablation_num_objects,tabl:FSC147_ablation_construction,tabl:FSC147_ablation_num_exemplars}.

\textbf{Keeping backbone frozen works best.} 
In~\Cref{tabl:FSC147_ablation_model_arch}, we evaluate the effect of unfreezing the last two backbone blocks shared by $\Phi$ and $\Psi$. In addition, we test training a CNN-based encoder $\Psi$ from scratch similar to \citet{liu2022countr}.
We find that keeping both frozen, \ie $\Phi=\Psi=\text{const.}$ works best, as it discourages the visual features to adapt to potential artefacts in our Self-Collages.

\textbf{Benefit of self-supervised exemplars.}
In~\Cref{tabl:FSC147_ablation_num_exemplars} we evaluate the effect of including the zero-shot counting task (\ie counting all salient objects) as done in other works~\citep{liu2022countr,djukic2022low}.
However, we find that including this task even with only a 25\% chance leads to a lower performance.
This is likely due to the zero-shot task overfitting our Self-Collages and finding short-cuts, such as pasting artefacts, as we relax the constraint introduced in \Cref{sec:construction_strategy} and vary the number of pasted objects to match the desired target. 
We therefore train our model with 1-3 exemplars and show how we can still conduct semantic zero-shot counting in \Cref{sec:zero-shot}.

\begin{table}[tb]
    \vspace{-4mm}
    \centering
    \small
    \setlength{\tabcolsep}{0.2em}
    \begin{tabular}{l  ll c  ll }
        \toprule
        \multirow{2}*{Methods} & \multicolumn{2}{c}{Val}  & & \multicolumn{2}{c}{Test}\\
        \cmidrule{2-3} \cmidrule{5-6}
        & MAE$\downarrow$ & RMSE$\downarrow$ && MAE$\downarrow$ & RMSE$\downarrow$\\
        \midrule
        CounTR \citep{liu2022countr}   & 13.13 & 49.83 && 11.95 & 91.23 \\
        LOCA \citep{djukic2022low}     & \textbf{10.24} & \textbf{32.56} && \textbf{10.79} & \textbf{56.97}\\
        \midrule
        \textbf{UnCounTR (ours)}          & 36.93 & 106.61 &&  35.77 & 130.34 \\
        \multicolumn{1}{r}{$\sigma$(5 runs)} & ${\pm0.49}$ & ${\pm1.46}$ && ${\pm0.60}$ & ${\pm0.94}$ \\
        \bottomrule
    \end{tabular}
    \vspace{-2mm}
    \caption{\textbf{Comparison to supervised models.} We evaluate on the val and test split of FSC-147.}
    \label{tabl:FSC147_comparison}
    \vspace{-5mm}
\end{table}

\begin{table*}
    \centering
    \small
    \vspace{-1mm}
    \setlength{\tabcolsep}{0.3em}
    \begin{tabular}{l @{\hskip 0.5em} rrll c rrll c rrll }
        \toprule
        \multirow{2}*{Method}  & \multicolumn{4}{c}{FSC-147 \textit{low} (8-16 objects)}  && \multicolumn{4}{c}{FSC-147 \textit{medium} (17-40 objects)}  && \multicolumn{4}{c}{FSC-147 \textit{high} (41-3701 objects)} \\
        \cmidrule{2-5} \cmidrule{7-10} \cmidrule{12-15}
        & MAE$\downarrow$ & RMSE$\downarrow$ & $\Delta$ & $\tau\uparrow$ && MAE$\downarrow$ & RMSE$\downarrow$ & $\Delta$ & $\tau\uparrow$ && MAE$\downarrow$ & RMSE$\downarrow$ & $\Delta$ & $\tau\uparrow$\\
        \midrule
        CounTR & 6.58 & 48.11 && \textbf{0.42} && \textbf{4.48} & \textbf{10.02} && \textbf{0.60} &&
        \textbf{20.50} & \textbf{126.95} && \textbf{0.80}\\
        \midrule
        \textbf{UnCounTR (ours)}  & \textbf{5.60} & \textbf{10.13} &{\color{ForestGreen}-37.9} &0.27&& 9.48 & 12.73  &{\color{red}+2.7} & 0.34 && 67.17 & {189.76} & {\color{red}+62.8} & {0.26}\\
        \multicolumn{1}{r}{$\sigma\text{(5 runs)}$}  & $\pm0.48$& $\pm0.84$ & &$\pm0.02$&& $\pm0.19$& $\pm0.33$  & & $\pm0.02$&& $\pm1.03$&  $\pm1.38$  & & $\pm0.01$\\
        \bottomrule
    \end{tabular}
    \vspace{-2mm}
    \caption{\textbf{Comparison to CounTR.} Evaluation results on the FSC-147 \textit{low} and \textit{medium} test subsets.}
    \label{tabl:FSC147_countr_comparison}
    \vspace{-2mm}

\vspace{-3mm}
\end{table*}
\textbf{Maximum number of pasted objects.}
Next, we evaluate the effect of varying $n_\text{max}$, the maximum number of objects pasted, as shown in~\Cref{tabl:FSC147_ablation_num_objects}.
Pasting up to 50 objects yields overall better performance on the full FSC-147 test dataset which contains on average 66 objects. While this shows that the construction of Self-Collages can be successfully scaled to higher counts, we find that pasting with 20 objects already achieves good performance with shortened construction times and so use this setting.

\textbf{Segmented pasting works best.}
From~\Cref{tabl:FSC147_ablation_construction}, We find that the best construction strategy involves pasting the self-supervised segmentations, regardless of their overlap with other objects. 
We simply store the segmentations and combine these to create diverse Self-Collages at training time.

\textbf{Compatible with various pretrained backbones.} 
In~\Cref{tabl:FSC147_ablation_backbone}, we show the effect of using different frozen weights for the visual encoder. 
Across ViT-B models with varying patch sizes and pretrainings, from Leopart's~\citep{ziegler2022self} spatially dense to the original DINO's~\citep{caron2021emerging} image-level pretraining, we find similar performances across our evaluations.
This shows that our method is compatible with various architectures and pretrainings and will likely benefit from the continued progress in self-supervised representation learning which is supported by our findings based on DINOv2~\cite{oquab2023dinov2} presented in \Cref{sec:improvements}.
We do not find any benefit in going from patch size 16 to 8 (similar to~\citet{LOST}) so we use the faster DINO ViT-B/16 model for the rest of the paper.

\begin{table}%
    \vspace{-1mm}
    \centering
    \small
    \setlength{\tabcolsep}{0.4em}
    \begin{tabular}{l @{\hskip 0.5em} ccc }
        \toprule
        \multirow{2}*{Method}  & \multicolumn{3}{c}{MSO} \\
        \cmidrule{2-4}
        & MAE$\downarrow$ & RMSE$\downarrow$ & $\tau\uparrow$\\
        \midrule
        Conn. Comp. Baseline & 10.67 & 12.88 & 0.12 \\
        CounTR & {2.34} & 8.12 & 0.36 \\
        \midrule
        \textbf{UnCounTR (ours)}  & {1.07} & {2.32} & {0.45} \\
        \multicolumn{1}{r}{$\sigma\text{(5 runs)}$}  & $\pm 0.06$ & $\pm 0.25$ & $\pm 0.01$ \\
        \bottomrule
    \end{tabular}
    \vspace{-2mm}
    \caption{\textbf{Evaluation on the MSO \cite{zhang2015salient} dataset}. %
    \label{tabl:MSO}}
    \vspace{-6mm}
\end{table}

\subsection{Benchmark comparison}

Next, we compare with previous methods on three datasets: FSC-147~\citep{ranjan2021learning}, MSO~\citep{zhang2015salient}, and CARPK~\citep{hsieh2017drone}. 
\Cref{tabl:FSC147_comparison} shows the results on the validation and test split of the FSC-147 dataset.
In contrast to our approach, which investigates the feasibility of learning to count without any reliance on supervised data, these methods heavily depend on manual annotations.
Hence, the comparison is certainly not fair: (1) our method does not use manually annotated counting data, also has never seen any FSC-147 training images, (2) our training samples based on Self-Collages only cover 3-19 target objects, whereas the full test set of FSC-147 has 66 objects on average.
Evaluating our method on the full set of FSC-147 is actually evaluating the transferring ability as discussed in more detail in \Cref{sec:generatlisation_ood}. 

We choose CounTR~\citep{liu2022countr} for further analysis because of its similar architecture to UnCounTR. 
To fairly compare with this method,
we trisect the FSC-147 test set as described in~\Cref{sec:implementation_details}.
\Cref{tabl:FSC147_countr_comparison} shows the evaluation results on the three partitions.
It is remarkable that our method outperforms supervised CounTR on the \textit{low} partition, especially on the RMSE metric (10.13 vs 48.11).
Also, our method is not far from CounTR's performance on the \textit{medium} partition (\eg~RMSE 12.7 vs 10.02), showing a reasonable transferring ability since the model is trained with up to 19 object counts only.

\begin{figure*}
    \centering
    \includegraphics[width=0.85\textwidth]{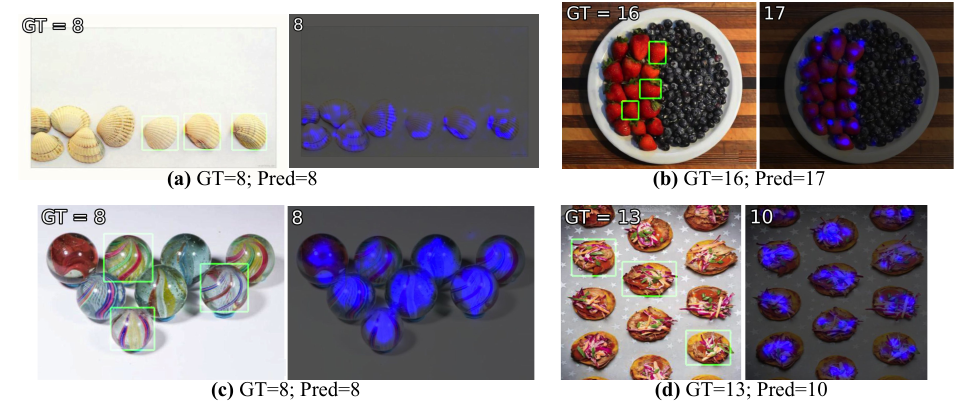}
    \vspace{-3mm}
    \caption{\textbf{Qualitative examples of UnCounTR's predictions.} We show predictions on four images from the FSC-147 test set, the green boxes represent the exemplars. Our predicted count is the sum of the density map rounded to the nearest integer.}
    \label{fig:pred_qualitative}
    \vspace{-1mm}
\end{figure*}
\begin{figure*}[ht]
    \centering
    \includegraphics[width=\textwidth]{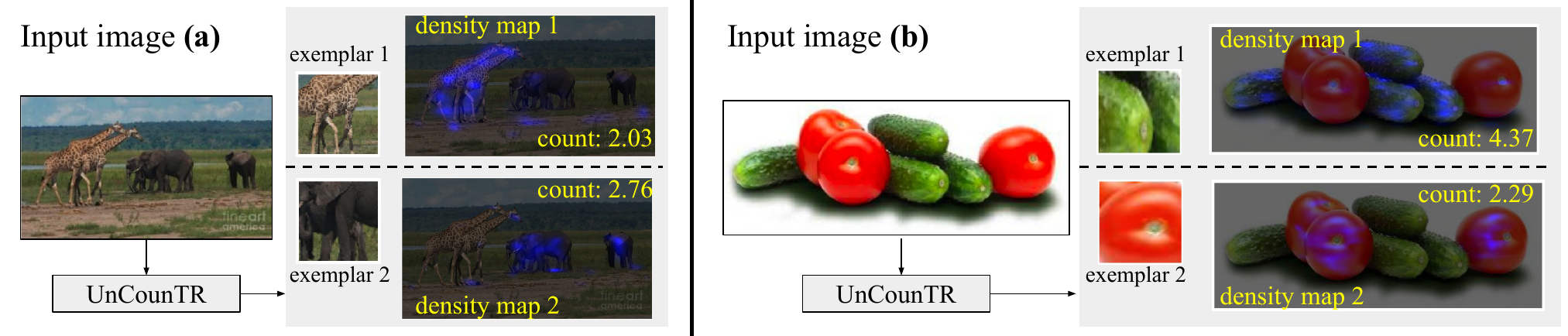}
    \vspace{-6mm}
    \caption{\textbf{Self-supervised semantic counting.} In this setting, the model proposes the exemplars by itself and then performs reference-based counting.}
    \label{fig:self_exemplar}
    \vspace{-4mm}
\end{figure*}

Additionally, \Cref{tabl:MSO} compares our method with the connected components baseline and CounTR on a subset of MSO as described in~\Cref{sec:implementation_details}.
The results show our self-supervised method outperforms the supervised CounTR model by a large margin on the lower counts tasks.
In \Cref{tabl:CARPK}, the performance on the CARPK dataset is shown.
We use the same CounTR model finetuned on the FSC-147 dataset as before to compare the generalisation across datasets. While the MAE of CounTR almost doubles compared to the results on the FSC-147 test set, UnCounTR's performance is more stable, improving by 15\%.
\begin{table}[t]
    \vspace{-1mm}
    \centering
    \small
    \setlength{\tabcolsep}{0.4em}
    \begin{tabular}{l @{\hskip 0.5em} ccc }
        \toprule
        \multirow{2}*{Method}  & \multicolumn{3}{c}{CARPK} \\
        \cmidrule{2-4}
        & MAE$\downarrow$ & RMSE$\downarrow$ & $\tau\uparrow$\\
        \midrule
        Conn. Comp. Baseline & 49.62 & 54.50 & 0.56 \\
        CounTR & {19.62} & 29.70 & 0.57 \\
        \midrule
        \textbf{UnCounTR (ours)}  & {30.35} & {35.67} & {0.54} \\
        \multicolumn{1}{r}{$\sigma\text{(5 runs)}$}  & $\pm 2.26$ & $\pm 2.36$ & $\pm 0.06$\\
        \bottomrule
    \end{tabular}
    \vspace{-2mm}
    \caption{\textbf{Evaluation on the CARPK \cite{hsieh2017drone} dataset}. 
    \label{tabl:CARPK}}
    \vspace{-6mm}
\end{table}

\subsection{Qualitative results and limitations \label{sec:qualitative}}

\Cref{fig:pred_qualitative} shows a few qualitative results to demonstrate our method's effectiveness and its limitations.
The model is able to correctly predict the number of objects for clearly separated and slightly overlapping instances (\eg~Subfigures~\textbf{a} and \textbf{c}).
The model also successfully identifies the object type of interest, \eg in Subfigure~\textbf{b} the density map correctly highlights the strawberries rather than blueberries.
One limitation of our model is images where the number of objects is ambiguous. For example, in Subfigure~\textbf{d}, the prediction missed a few burgers which are possibly the ones partially shown on the edge. 
However, it is also difficult for humans to decide which of those objects should be counted.

\subsection{Improving upon UnCounTR}
Having established UnCounTR and compared it to several methods, we now explore three ways to further improve its performance: (1) Integrating the DINOv2~\citep{oquab2023dinov2} backbone improves the MAE particularly on the FSC-147 \textit{low} set by 27\% (2) Using cluster similarity to create more challenging Self-Collages reduces the MAE on FSC-147 to 31.06 (3) Refining high-count predictions further improves performance yielding the final \textbf{UnCounTRv2} model. It achieves an MAE of 28.67 on FSC-147 and 3.88 on FSC-147 \textit{low}. For a comprehensive discussion, please refer to \Cref{sec:improvements}. See \Cref{fig:uncountrv2_countr} for a visual comparison with CounTR.

\subsection{Self-supervised semantic counting \label{sec:zero-shot}}
In this last section, we explore the potential of UnCounTR for more advanced counting tasks. 
In particular, we test whether our model can not only unsupervisedly count different kinds of objects in an image, but also determine the categories by itself, a scenario we refer to as \textit{semantic} counting.
To this end, we use a simple pipeline that picks an area surrounding the maximum in the input image's \texttt{CLS}-attention map as the first input and refines it to obtain an exemplar.
Next, UnCounTR predicts the number of objects in the image based on the self-constructed exemplar.
Finally, the locations that have been detected using this procedure are removed from the initial attention map and if the remaining attention values are high enough, the process is repeated to count objects of another type. We provide the details of this method in \Cref{sec:self_supervised_counting_details}.
In~\Cref{fig:self_exemplar}, we demonstrate results on two difficult real images.

\section{Conclusion}
In this work, we have introduced UnCounTR, showcasing, for the first time, the capability of reference-based counting without relying on human supervision.
To this end, our method constructs Self-Collages, a simple unsupervised way of generating proxy learning signals from unlabeled data.
Our results demonstrate that by utilising an off-the-shelf unsupervisedly pretrained visual encoder, we can learn counting models that can even outperform strong baselines such as DETR and achieve similar performances to dedicated counting models such as CounTR on CARPK, MSO, and various splits of FSC-147.
Finally, we have shown that our model can unsupervisedly identify multiple exemplars in an image and count them, something no supervised model can yet do.
This work opens a wide space for extending unsupervised visual understanding beyond simple image-level representations to more complex tasks previously out of reach, such as scene graphs or unsupervised semantic instance segmentations.

\section*{Acknowledgement}
We would like to thank the ELLIS unit Amsterdam for funding LK's academic visit to VGG at the University of Oxford as part of the MSc Honours Programme.
TH would like to thank Niki Amini-Naieni for the helpful discussions.
LK thanks SURF (\href{https://www.surf.nl/en}{www.surf.nl}) for the support in using the National Supercomputer Snellius.
{
    \small
    \bibliographystyle{ieeenat_fullname}
    \bibliography{main}

\begin{thebibliography}{75}
\providecommand{\natexlab}[1]{#1}
\providecommand{\url}[1]{\texttt{#1}}
\expandafter\ifx\csname urlstyle\endcsname\relax
  \providecommand{\doi}[1]{doi: #1}\else
  \providecommand{\doi}{doi: \begingroup \urlstyle{rm}\Url}\fi

\bibitem[Alayrac et~al.(2020)Alayrac, Recasens, Schneider, Arandjelovi{\'c}, Ramapuram, De~Fauw, Smaira, Dieleman, and Zisserman]{alayrac2020mmv}
Jean-Baptiste Alayrac, Adria Recasens, Rosalia Schneider, Relja Arandjelovi{\'c}, Jason Ramapuram, Jeffrey De~Fauw, Lucas Smaira, Sander Dieleman, and Andrew Zisserman.
\newblock Self-supervised multimodal versatile networks.
\newblock \emph{NeurIPS}, 33:\penalty0 25--37, 2020.

\bibitem[Arandjelovi{\'c} and Zisserman(2019)]{arandjelovic2019object}
Relja Arandjelovi{\'c} and Andrew Zisserman.
\newblock Object discovery with a copy-pasting gan.
\newblock \emph{arXiv preprint arXiv:1905.11369}, 2019.

\bibitem[Asano et~al.(2020)Asano, Rupprecht, and Vedaldi]{asano2019self}
Yuki~Markus Asano, Christian Rupprecht, and Andrea Vedaldi.
\newblock Self-labelling via simultaneous clustering and representation learning.
\newblock \emph{ICLR}, 2020.

\bibitem[Baradad~Jurjo et~al.(2021)Baradad~Jurjo, Wulff, Wang, Isola, and Torralba]{baradad2021learning}
Manel Baradad~Jurjo, Jonas Wulff, Tongzhou Wang, Phillip Isola, and Antonio Torralba.
\newblock Learning to see by looking at noise.
\newblock \emph{Advances in Neural Information Processing Systems}, 34:\penalty0 2556--2569, 2021.

\bibitem[Buciluǎ et~al.(2006)Buciluǎ, Caruana, and Niculescu-Mizil]{bucilua2006model}
Cristian Buciluǎ, Rich Caruana, and Alexandru Niculescu-Mizil.
\newblock Model compression.
\newblock In \emph{Proceedings of the 12th ACM SIGKDD international conference on Knowledge discovery and data mining}, pages 535--541, 2006.

\bibitem[Carion et~al.(2020)Carion, Massa, Synnaeve, Usunier, Kirillov, and Zagoruyko]{carion2020end}
Nicolas Carion, Francisco Massa, Gabriel Synnaeve, Nicolas Usunier, Alexander Kirillov, and Sergey Zagoruyko.
\newblock End-to-end object detection with transformers.
\newblock In \emph{ECCV}, pages 213--229. Springer, 2020.

\bibitem[Caron et~al.(2018)Caron, Bojanowski, Joulin, and Douze]{caron2018deep}
Mathilde Caron, Piotr Bojanowski, Armand Joulin, and Matthijs Douze.
\newblock Deep clustering for unsupervised learning of visual features.
\newblock In \emph{ECCV}, pages 132--149, 2018.

\bibitem[Caron et~al.(2021)Caron, Touvron, Misra, J{\'e}gou, Mairal, Bojanowski, and Joulin]{caron2021emerging}
Mathilde Caron, Hugo Touvron, Ishan Misra, Herv{\'e} J{\'e}gou, Julien Mairal, Piotr Bojanowski, and Armand Joulin.
\newblock Emerging properties in self-supervised vision transformers.
\newblock In \emph{ICCV}, pages 9650--9660, 2021.

\bibitem[Chen et~al.(2017)Chen, Choi, Yu, Han, and Chandraker]{chen2017learning}
Guobin Chen, Wongun Choi, Xiang Yu, Tony Han, and Manmohan Chandraker.
\newblock Learning efficient object detection models with knowledge distillation.
\newblock \emph{NeurIPS}, 30, 2017.

\bibitem[Chen et~al.(2022)Chen, Zhou, Li, Wei, and Xiao]{chen2022self}
Hao Chen, Yangzhun Zhou, Jun Li, Xiu-Shen Wei, and Liang Xiao.
\newblock Self-supervised multi-category counting networks for automatic check-out.
\newblock \emph{IEEE Transactions on Image Processing}, 31:\penalty0 3004--3016, 2022.

\bibitem[Chen et~al.(2020)Chen, Kornblith, Norouzi, and Hinton]{chen2020simple}
Ting Chen, Simon Kornblith, Mohammad Norouzi, and Geoffrey Hinton.
\newblock A simple framework for contrastive learning of visual representations.
\newblock In \emph{ICML}, pages 1597--1607. PMLR, 2020.

\bibitem[Davis and P{\'e}russe(1988)]{davis1988numerical}
Hank Davis and Rachelle P{\'e}russe.
\newblock Numerical competence in animals: Definitional issues, current evidence, and a new research agenda.
\newblock \emph{Behavioral and Brain Sciences}, 11\penalty0 (4):\penalty0 561--579, 1988.

\bibitem[Dehaene(2011)]{dehaene2011number}
Stanislas Dehaene.
\newblock \emph{The number sense: How the mind creates mathematics}.
\newblock Oxford University Press, 2011.

\bibitem[Deng et~al.(2009)Deng, Dong, Socher, Li, Li, and Fei-Fei]{deng2009imagenet}
Jia Deng, Wei Dong, Richard Socher, Li-Jia Li, Kai Li, and Li Fei-Fei.
\newblock Imagenet: A large-scale hierarchical image database.
\newblock In \emph{CVPR}, pages 248--255. Ieee, 2009.

\bibitem[Djukic et~al.(2022)Djukic, Lukezic, Zavrtanik, and Kristan]{djukic2022low}
Nikola Djukic, Alan Lukezic, Vitjan Zavrtanik, and Matej Kristan.
\newblock A low-shot object counting network with iterative prototype adaptation.
\newblock \emph{arXiv preprint arXiv:2211.08217}, 2022.

\bibitem[Doersch et~al.(2015)Doersch, Gupta, and Efros]{doersch2015unsupervised}
Carl Doersch, Abhinav Gupta, and Alexei~A Efros.
\newblock Unsupervised visual representation learning by context prediction.
\newblock In \emph{ICCV}, pages 1422--1430, 2015.

\bibitem[Dosovitskiy et~al.(2021)Dosovitskiy, Beyer, Kolesnikov, Weissenborn, Zhai, Unterthiner, Dehghani, Minderer, Heigold, Gelly, et~al.]{dosovitskiy2020image}
Alexey Dosovitskiy, Lucas Beyer, Alexander Kolesnikov, Dirk Weissenborn, Xiaohua Zhai, Thomas Unterthiner, Mostafa Dehghani, Matthias Minderer, Georg Heigold, Sylvain Gelly, et~al.
\newblock An image is worth 16x16 words: Transformers for image recognition at scale.
\newblock \emph{ICLR}, 2021.

\bibitem[Dvornik et~al.(2018)Dvornik, Mairal, and Schmid]{dvornik2018modeling}
Nikita Dvornik, Julien Mairal, and Cordelia Schmid.
\newblock Modeling visual context is key to augmenting object detection datasets.
\newblock In \emph{Proceedings of the European Conference on Computer Vision (ECCV)}, pages 364--380, 2018.

\bibitem[Ge et~al.(2021)Ge, Mishra, Li, Wang, and Jacobs]{ge2021robust}
Songwei Ge, Shlok Mishra, Chun-Liang Li, Haohan Wang, and David Jacobs.
\newblock Robust contrastive learning using negative samples with diminished semantics.
\newblock \emph{Advances in Neural Information Processing Systems}, 34:\penalty0 27356--27368, 2021.

\bibitem[Ghiasi et~al.(2021)Ghiasi, Cui, Srinivas, Qian, Lin, Cubuk, Le, and Zoph]{ghiasi2021simple}
Golnaz Ghiasi, Yin Cui, Aravind Srinivas, Rui Qian, Tsung-Yi Lin, Ekin~D Cubuk, Quoc~V Le, and Barret Zoph.
\newblock Simple copy-paste is a strong data augmentation method for instance segmentation.
\newblock In \emph{Proceedings of the IEEE/CVF conference on computer vision and pattern recognition}, pages 2918--2928, 2021.

\bibitem[Grill et~al.(2020)Grill, Strub, Altch{\'e}, Tallec, Richemond, Buchatskaya, Doersch, Avila~Pires, Guo, Gheshlaghi~Azar, et~al.]{grill2020bootstrap}
Jean-Bastien Grill, Florian Strub, Florent Altch{\'e}, Corentin Tallec, Pierre Richemond, Elena Buchatskaya, Carl Doersch, Bernardo Avila~Pires, Zhaohan Guo, Mohammad Gheshlaghi~Azar, et~al.
\newblock Bootstrap your own latent-a new approach to self-supervised learning.
\newblock \emph{NeurIPS}, 33:\penalty0 21271--21284, 2020.

\bibitem[Han et~al.(2019)Han, Xie, and Zisserman]{han2019video}
Tengda Han, Weidi Xie, and Andrew Zisserman.
\newblock Video representation learning by dense predictive coding.
\newblock In \emph{ICCV Workshops}, pages 0--0, 2019.

\bibitem[He et~al.(2016)He, Zhang, Ren, and Sun]{he2016deep}
Kaiming He, Xiangyu Zhang, Shaoqing Ren, and Jian Sun.
\newblock Deep residual learning for image recognition.
\newblock In \emph{CVPR}, pages 770--778, 2016.

\bibitem[He et~al.(2020)He, Fan, Wu, Xie, and Girshick]{he2020momentum}
Kaiming He, Haoqi Fan, Yuxin Wu, Saining Xie, and Ross Girshick.
\newblock Momentum contrast for unsupervised visual representation learning.
\newblock In \emph{CVPR}, pages 9729--9738, 2020.

\bibitem[He et~al.(2022)He, Chen, Xie, Li, Doll{\'a}r, and Girshick]{he2022masked}
Kaiming He, Xinlei Chen, Saining Xie, Yanghao Li, Piotr Doll{\'a}r, and Ross Girshick.
\newblock Masked autoencoders are scalable vision learners.
\newblock In \emph{CVPR}, pages 16000--16009, 2022.

\bibitem[Hobley and Prisacariu(2022)]{hobley2022learning}
Michael Hobley and Victor Prisacariu.
\newblock Learning to count anything: Reference-less class-agnostic counting with weak supervision.
\newblock \emph{arXiv preprint arXiv:2205.10203}, 2022.

\bibitem[Hsieh et~al.(2017)Hsieh, Lin, and Hsu]{hsieh2017drone}
Meng-Ru Hsieh, Yen-Liang Lin, and Winston~H Hsu.
\newblock Drone-based object counting by spatially regularized regional proposal network.
\newblock In \emph{ICCV}, pages 4145--4153, 2017.

\bibitem[Karras et~al.(2020)Karras, Laine, Aittala, Hellsten, Lehtinen, and Aila]{karras2020analyzing}
Tero Karras, Samuli Laine, Miika Aittala, Janne Hellsten, Jaakko Lehtinen, and Timo Aila.
\newblock Analyzing and improving the image quality of stylegan.
\newblock In \emph{Proceedings of the IEEE/CVF conference on computer vision and pattern recognition}, pages 8110--8119, 2020.

\bibitem[Katsuki et~al.(2016)Katsuki, Morimura, and Id{\'e}]{katsuki2016unsupervised}
Takayuki Katsuki, Tetsuro Morimura, and Tsuyoshi Id{\'e}.
\newblock Unsupervised object counting without object recognition.
\newblock In \emph{2016 23rd International Conference on Pattern Recognition (ICPR)}, pages 3627--3632. IEEE, 2016.

\bibitem[Kim and Rush(2016)]{kim2016sequence}
Yoon Kim and Alexander~M Rush.
\newblock Sequence-level knowledge distillation.
\newblock \emph{arXiv preprint arXiv:1606.07947}, 2016.

\bibitem[Kirillov et~al.(2023)Kirillov, Mintun, Ravi, Mao, Rolland, Gustafson, Xiao, Whitehead, Berg, Lo, et~al.]{kirillov2023sam}
Alexander Kirillov, Eric Mintun, Nikhila Ravi, Hanzi Mao, Chloe Rolland, Laura Gustafson, Tete Xiao, Spencer Whitehead, Alexander~C Berg, Wan-Yen Lo, et~al.
\newblock Segment anything.
\newblock \emph{arXiv preprint arXiv:2304.02643}, 2023.

\bibitem[Krizhevsky et~al.(2012)Krizhevsky, Sutskever, and Hinton]{Krizhevsky2012alexnet}
Alex Krizhevsky, Ilya Sutskever, and Geoffrey~E Hinton.
\newblock Imagenet classification with deep convolutional neural networks.
\newblock In \emph{NeurIPS}, 2012.

\bibitem[Liang et~al.(2023)Liang, Xie, Zou, Ye, Xu, and Bai]{liang2023crowdclip}
Dingkang Liang, Jiahao Xie, Zhikang Zou, Xiaoqing Ye, Wei Xu, and Xiang Bai.
\newblock Crowdclip: Unsupervised crowd counting via vision-language model.
\newblock In \emph{Proceedings of the IEEE/CVF Conference on Computer Vision and Pattern Recognition}, pages 2893--2903, 2023.

\bibitem[Lin et~al.(2021)Lin, Hong, and Wang]{lin2021object}
Hui Lin, Xiaopeng Hong, and Yabin Wang.
\newblock Object counting: You only need to look at one.
\newblock \emph{arXiv preprint arXiv:2112.05993}, 2021.

\bibitem[Lin et~al.(2014)Lin, Maire, Belongie, Hays, Perona, Ramanan, Doll{\'a}r, and Zitnick]{lin2014microsoft}
Tsung-Yi Lin, Michael Maire, Serge Belongie, James Hays, Pietro Perona, Deva Ramanan, Piotr Doll{\'a}r, and C~Lawrence Zitnick.
\newblock Microsoft coco: Common objects in context.
\newblock In \emph{ECCV}, pages 740--755. Springer, 2014.

\bibitem[Liu et~al.(2022)Liu, Zhong, Zisserman, and Xie]{liu2022countr}
Chang Liu, Yujie Zhong, Andrew Zisserman, and Weidi Xie.
\newblock Countr: Transformer-based generalised visual counting.
\newblock \emph{BMVC}, 2022.

\bibitem[Liu et~al.(2018)Liu, Van De~Weijer, and Bagdanov]{liu2018leveraging}
Xialei Liu, Joost Van De~Weijer, and Andrew~D Bagdanov.
\newblock Leveraging unlabeled data for crowd counting by learning to rank.
\newblock In \emph{CVPR}, pages 7661--7669, 2018.

\bibitem[Lu et~al.(2018)Lu, Xie, and Zisserman]{lu2018class}
Erika Lu, Weidi Xie, and Andrew Zisserman.
\newblock Class-agnostic counting.
\newblock In \emph{ACCV}, pages 669--684. Springer, 2018.

\bibitem[Melas-Kyriazi et~al.(2022)Melas-Kyriazi, Rupprecht, Laina, and Vedaldi]{melas2022deep}
Luke Melas-Kyriazi, Christian Rupprecht, Iro Laina, and Andrea Vedaldi.
\newblock Deep spectral methods: A surprisingly strong baseline for unsupervised semantic segmentation and localization.
\newblock In \emph{CVPR}, pages 8364--8375, 2022.

\bibitem[Misra et~al.(2016)Misra, Zitnick, and Hebert]{misra2016shuffle}
Ishan Misra, C~Lawrence Zitnick, and Martial Hebert.
\newblock Shuffle and learn: unsupervised learning using temporal order verification.
\newblock In \emph{ECCV}, pages 527--544. Springer, 2016.

\bibitem[Mundhenk et~al.(2016)Mundhenk, Konjevod, Sakla, and Boakye]{mundhenk2016large}
T~Nathan Mundhenk, Goran Konjevod, Wesam~A Sakla, and Kofi Boakye.
\newblock A large contextual dataset for classification, detection and counting of cars with deep learning.
\newblock In \emph{ECCV}, pages 785--800. Springer, 2016.

\bibitem[Noroozi and Favaro(2016)]{noroozi2016jigsaw}
Mehdi Noroozi and Paolo Favaro.
\newblock Unsupervised learning of visual representations by solving jigsaw puzzles.
\newblock In \emph{ECCV}, pages 69--84. Springer, 2016.

\bibitem[Noroozi et~al.(2017)Noroozi, Pirsiavash, and Favaro]{noroozi2017representation}
Mehdi Noroozi, Hamed Pirsiavash, and Paolo Favaro.
\newblock Representation learning by learning to count.
\newblock In \emph{ICCV}, pages 5898--5906, 2017.

\bibitem[Oord et~al.(2018)Oord, Li, and Vinyals]{oord2018representation}
Aaron van~den Oord, Yazhe Li, and Oriol Vinyals.
\newblock Representation learning with contrastive predictive coding.
\newblock \emph{arXiv preprint arXiv:1807.03748}, 2018.

\bibitem[Oquab et~al.(2023)Oquab, Darcet, Moutakanni, Vo, Szafraniec, Khalidov, Fernandez, Haziza, Massa, El-Nouby, et~al.]{oquab2023dinov2}
Maxime Oquab, Timoth{\'e}e Darcet, Th{\'e}o Moutakanni, Huy Vo, Marc Szafraniec, Vasil Khalidov, Pierre Fernandez, Daniel Haziza, Francisco Massa, Alaaeldin El-Nouby, et~al.
\newblock Dinov2: Learning robust visual features without supervision.
\newblock \emph{arXiv preprint arXiv:2304.07193}, 2023.

\bibitem[Pahl et~al.(2013)Pahl, Si, and Zhang]{pahl2013numerical}
Mario Pahl, Aung Si, and Shaowu Zhang.
\newblock Numerical cognition in bees and other insects.
\newblock \emph{Frontiers in psychology}, 4:\penalty0 162, 2013.

\bibitem[Pathak et~al.(2016)Pathak, Krahenbuhl, Donahue, Darrell, and Efros]{pathak2016context}
Deepak Pathak, Philipp Krahenbuhl, Jeff Donahue, Trevor Darrell, and Alexei~A Efros.
\newblock Context encoders: Feature learning by inpainting.
\newblock In \emph{CVPR}, pages 2536--2544, 2016.

\bibitem[Piazza and Dehaene(2004)]{piazza2004number}
Manuela Piazza and Stanislas Dehaene.
\newblock From number neurons to mental arithmetic: The cognitive neuroscience of number sense.
\newblock \emph{The cognitive neurosciences, 3rd edition, ed. MS Gazzaniga}, pages 865--77, 2004.

\bibitem[Radford et~al.(2021)Radford, Kim, Hallacy, Ramesh, Goh, Agarwal, Sastry, Askell, Mishkin, Clark, et~al.]{radford2021clip}
Alec Radford, Jong~Wook Kim, Chris Hallacy, Aditya Ramesh, Gabriel Goh, Sandhini Agarwal, Girish Sastry, Amanda Askell, Pamela Mishkin, Jack Clark, et~al.
\newblock Learning transferable visual models from natural language supervision.
\newblock In \emph{ICML}, pages 8748--8763. PMLR, 2021.

\bibitem[Ranjan et~al.(2021)Ranjan, Sharma, Nguyen, and Hoai]{ranjan2021learning}
Viresh Ranjan, Udbhav Sharma, Thu Nguyen, and Minh Hoai.
\newblock Learning to count everything.
\newblock In \emph{CVPR}, pages 3394--3403, 2021.

\bibitem[Ren et~al.(2015)Ren, He, Girshick, and Sun]{ren2015faster}
Shaoqing Ren, Kaiming He, Ross Girshick, and Jian Sun.
\newblock Faster r-cnn: Towards real-time object detection with region proposal networks.
\newblock \emph{NeurIPS}, 28, 2015.

\bibitem[Russakovsky et~al.(2015)Russakovsky, Deng, Su, Krause, Satheesh, Ma, Huang, Karpathy, Khosla, Bernstein, et~al.]{russakovsky2015imagenet}
Olga Russakovsky, Jia Deng, Hao Su, Jonathan Krause, Sanjeev Satheesh, Sean Ma, Zhiheng Huang, Andrej Karpathy, Aditya Khosla, Michael Bernstein, et~al.
\newblock Imagenet large scale visual recognition challenge.
\newblock \emph{International journal of computer vision}, 115:\penalty0 211--252, 2015.

\bibitem[Sam et~al.(2019)Sam, Sajjan, Maurya, and Babu]{sam2019almost}
Deepak~Babu Sam, Neeraj~N Sajjan, Himanshu Maurya, and R~Venkatesh Babu.
\newblock Almost unsupervised learning for dense crowd counting.
\newblock In \emph{Proceedings of the AAAI conference on artificial intelligence}, pages 8868--8875, 2019.

\bibitem[Sam et~al.(2020)Sam, Agarwalla, Joseph, Sindagi, Babu, and Patel]{sam2020completely}
Deepak~Babu Sam, Abhinav Agarwalla, Jimmy Joseph, Vishwanath~A Sindagi, R~Venkatesh Babu, and Vishal~M Patel.
\newblock Completely self-supervised crowd counting via distribution matching.
\newblock \emph{arXiv preprint arXiv:2009.06420}, 2020.

\bibitem[Shi et~al.(2019)Shi, Mettes, and Snoek]{shi2019counting}
Zenglin Shi, Pascal Mettes, and Cees~GM Snoek.
\newblock Counting with focus for free.
\newblock In \emph{Proceedings of the IEEE/CVF International Conference on Computer Vision}, pages 4200--4209, 2019.

\bibitem[Shi et~al.(2023)Shi, Sun, and Zhang]{shi2023training}
Zenglin Shi, Ying Sun, and Mengmi Zhang.
\newblock Training-free object counting with prompts.
\newblock \emph{arXiv preprint arXiv:2307.00038}, 2023.

\bibitem[Shin et~al.(2022)Shin, Albanie, and Xie]{shin2022unsupervised}
Gyungin Shin, Samuel Albanie, and Weidi Xie.
\newblock Unsupervised salient object detection with spectral cluster voting.
\newblock In \emph{CVPR}, pages 3971--3980, 2022.

\bibitem[Sim\'eoni et~al.(2021)Sim\'eoni, Puy, Vo, Roburin, Gidaris, Bursuc, P\'erez, Marlet, and Ponce]{LOST}
Oriane Sim\'eoni, Gilles Puy, Huy~V. Vo, Simon Roburin, Spyros Gidaris, Andrei Bursuc, Patrick P\'erez, Renaud Marlet, and Jean Ponce.
\newblock Localizing objects with self-supervised transformers and no labels.
\newblock In \emph{BMVC}, 2021.

\bibitem[Slaughter et~al.(2011)Slaughter, Itakura, Kutsuki, and Siegal]{slaughter2011learning}
Virginia Slaughter, Shoji Itakura, Aya Kutsuki, and Michael Siegal.
\newblock Learning to count begins in infancy: Evidence from 18 month olds' visual preferences.
\newblock \emph{Proceedings of the Royal Society B: Biological Sciences}, 278\penalty0 (1720):\penalty0 2979--2984, 2011.

\bibitem[Smith and Gasser(2005)]{smith2005development}
Linda Smith and Michael Gasser.
\newblock The development of embodied cognition: Six lessons from babies.
\newblock \emph{Artificial life}, 11\penalty0 (1-2):\penalty0 13--29, 2005.

\bibitem[Trick and Pylyshyn(1994)]{trick1994small}
Lana~M Trick and Zenon~W Pylyshyn.
\newblock Why are small and large numbers enumerated differently? a limited-capacity preattentive stage in vision.
\newblock \emph{Psychological review}, 101\penalty0 (1):\penalty0 80, 1994.

\bibitem[Ubbens et~al.(2020)Ubbens, Ayalew, Shirtliffe, Josuttes, Pozniak, and Stavness]{ubbens2020autocount}
Jordan~R Ubbens, Tewodros~W Ayalew, Steve Shirtliffe, Anique Josuttes, Curtis Pozniak, and Ian Stavness.
\newblock Autocount: Unsupervised segmentation and counting of organs in field images.
\newblock In \emph{Computer Vision--ECCV 2020 Workshops: Glasgow, UK, August 23--28, 2020, Proceedings, Part VI 16}, pages 391--399. Springer, 2020.

\bibitem[Van~Gansbeke et~al.(2020)Van~Gansbeke, Vandenhende, Georgoulis, Proesmans, and Van~Gool]{vangansbeke2020scan}
Wouter Van~Gansbeke, Simon Vandenhende, Stamatios Georgoulis, Marc Proesmans, and Luc Van~Gool.
\newblock Scan: Learning to classify images without labels.
\newblock In \emph{ECCV}, 2020.

\bibitem[Xiao et~al.(2010)Xiao, Hays, Ehinger, Oliva, and Torralba]{xiao2010sun}
Jianxiong Xiao, James Hays, Krista~A Ehinger, Aude Oliva, and Antonio Torralba.
\newblock Sun database: Large-scale scene recognition from abbey to zoo.
\newblock In \emph{CVPR}, pages 3485--3492. IEEE, 2010.

\bibitem[Xuan et~al.(2020)Xuan, Stylianou, Liu, and Pless]{xuan2020hard}
Hong Xuan, Abby Stylianou, Xiaotong Liu, and Robert Pless.
\newblock Hard negative examples are hard, but useful.
\newblock In \emph{Computer Vision--ECCV 2020: 16th European Conference, Glasgow, UK, August 23--28, 2020, Proceedings, Part XIV 16}, pages 126--142. Springer, 2020.

\bibitem[Yan et~al.(2020)Yan, Misra, Gupta, Ghadiyaram, and Mahajan]{yan2020clusterfit}
Xueting Yan, Ishan Misra, Abhinav Gupta, Deepti Ghadiyaram, and Dhruv Mahajan.
\newblock Clusterfit: Improving generalization of visual representations.
\newblock In \emph{CVPR}, pages 6509--6518, 2020.

\bibitem[Yang et~al.(2021)Yang, Su, Hsu, and Chen]{yang2021class}
Shuo-Diao Yang, Hung-Ting Su, Winston~H Hsu, and Wen-Chin Chen.
\newblock Class-agnostic few-shot object counting.
\newblock In \emph{Proceedings of the IEEE/CVF Winter Conference on Applications of Computer Vision}, pages 870--878, 2021.

\bibitem[You et~al.(2023)You, Yang, Luo, Lu, Cui, and Le]{you2023few}
Zhiyuan You, Kai Yang, Wenhan Luo, Xin Lu, Lei Cui, and Xinyi Le.
\newblock Few-shot object counting with similarity-aware feature enhancement.
\newblock In \emph{Proceedings of the IEEE/CVF Winter Conference on Applications of Computer Vision}, pages 6315--6324, 2023.

\bibitem[Zadaianchuk et~al.(2023)Zadaianchuk, Kleindessner, Zhu, Locatello, and Brox]{zadaianchuk2023unsupervised}
Andrii Zadaianchuk, Matthaeus Kleindessner, Yi Zhu, Francesco Locatello, and Thomas Brox.
\newblock Unsupervised semantic segmentation with self-supervised object-centric representations.
\newblock In \emph{ICLR}, 2023.

\bibitem[Zhang et~al.(2015)Zhang, Ma, Sameki, Sclaroff, Betke, Lin, Shen, Price, and Mech]{zhang2015salient}
Jianming Zhang, Shugao Ma, Mehrnoosh Sameki, Stan Sclaroff, Margrit Betke, Zhe Lin, Xiaohui Shen, Brian Price, and Radomir Mech.
\newblock Salient object subitizing.
\newblock In \emph{CVPR}, pages 4045--4054, 2015.

\bibitem[Zhang et~al.(2022)Zhang, Sohoni, Zhang, Finn, and Re]{zhang2022correct}
Michael Zhang, Nimit~S Sohoni, Hongyang~R Zhang, Chelsea Finn, and Christopher Re.
\newblock Correct-n-contrast: a contrastive approach for improving robustness to spurious correlations.
\newblock In \emph{International Conference on Machine Learning}, pages 26484--26516. PMLR, 2022.

\bibitem[Zhang et~al.(2016)Zhang, Isola, and Efros]{zhang2016colorful}
Richard Zhang, Phillip Isola, and Alexei~A Efros.
\newblock Colorful image colorization.
\newblock In \emph{ECCV}, pages 649--666. Springer, 2016.

\bibitem[Zhao et~al.(2022)Zhao, Sheng, Bao, Chen, Chen, Wen, Yuan, Liu, Zhou, Chu, et~al.]{zhao2022x}
Hanqing Zhao, Dianmo Sheng, Jianmin Bao, Dongdong Chen, Dong Chen, Fang Wen, Lu Yuan, Ce Liu, Wenbo Zhou, Qi Chu, et~al.
\newblock X-paste: Revisit copy-paste at scale with clip and stablediffusion.
\newblock \emph{arXiv preprint arXiv:2212.03863}, 2022.

\bibitem[Zhou et~al.(2022)Zhou, Girdhar, Joulin, Kr{\"a}henb{\"u}hl, and Misra]{zhou2022detecting}
Xingyi Zhou, Rohit Girdhar, Armand Joulin, Philipp Kr{\"a}henb{\"u}hl, and Ishan Misra.
\newblock Detecting twenty-thousand classes using image-level supervision.
\newblock In \emph{ECCV}, pages 350--368. Springer, 2022.

\bibitem[Ziegler and Asano(2022)]{ziegler2022self}
Adrian Ziegler and Yuki~M Asano.
\newblock Self-supervised learning of object parts for semantic segmentation.
\newblock In \emph{CVPR}, pages 14502--14511, 2022.

\end{thebibliography}
}
\onecolumn
\appendix

\section{Further implementation details}\label{sec:implementation_details_appendix}

\subsection{Model architecture}\label{sec:model_architecture_appendix}

\paragraph{Image encoder}
For the image encoder $\Phi$, we employ a ViT-B/16 pretrained using the DINO approach~\citep{caron2021emerging} as the backbone. It consists of 12 transformer blocks with 12 heads each and uses fixed, sinusoidal position embeddings. The transformer operates on $d=768$ dimensions and increases the hidden dimensions within the two-layer MLPs after each attention block by a factor of four to 3072 dimensions. Each linear layer in the MLP is followed by the GELU non-linearity. More information can be found in the original work from \citet{caron2021emerging}.
By default, we freeze all 86M parameters of the backbone.

\paragraph{Exemplar encoding}
To encode an exemplar $\textit{\textbf{E}}$ with fixed spatial dimensions $H'\times W'$ into a single feature vector $\textbf{z} \in \mathbb{R}^{d}$, we first pass it through the backbone of the exemplar encoder $\Psi$ to obtain a feature map ${\textbf{x}_\text{E}=\Psi(\textit{\textbf{E}})\in \mathbb{R}^{h \times w \times d}}$ where $\Psi=\Phi$. The final representation $\textbf{z}$ is derived by computing the weighted sum of $\textbf{x}_\text{E}$ across the spatial dimensions where the weight of each patch is determined by the attention in the final \texttt{CLS}-attention map of $\Psi$ averaged over the heads.

\paragraph{Feature interaction module} 
We follow the architecture proposed by \citet{liu2022countr} which uses 2 transformer blocks with 16 heads to modify the image embeddings by using self-attention. In addition to self-attention, each block utilises cross-attention where the keys and values are based on the encoded exemplars. Since this transformer operates on 512-dimensional feature vectors, $\textbf{x}$ and $\textbf{z}_j,\ j \in \{1,...,E\}$ are projected to these dimensions using a linear layer. To give the feature interaction module direct access to positional information, fixed, sinusoidal position embeddings are added to the feature map. Similar to the image encoder, the MLPs increase the dimensionality four times to 2048. This results in 8.8M parameters.

\paragraph{Decoder}
The decoder is built based on 4 convolutional layers that upscale the patch-level features to the original resolution to obtain the final density map.
It has a total of 3.0M parameters.
All blocks contain a convolutional layer with 256 output channels and a kernel size of $3\times 3$. Each of them is followed by group normalisation with 8 groups and a ReLU non-linearity. The last block has a final convolutional layer with $1\times 1$ filters that reduce the number of channels to 1 to match the desired output format. After each block, the spatial resolution is doubled using bilinear interpolation. The result is a density map $\mathbf{\hat{y}}\in\mathbb{R}^{H\times W}$ with the same resolution as the input image.

\subsection{Composer module details}\label{sec:composer_details}

We always place the images $\mathcal{I}_0$ of the target cluster  $c_0$ on top of the non-target images $\mathcal{I}_i,\ i \in \{1,...,n_c-1\}$. This reduces the noise of Self-Collages since target objects can only be occluded by other target objects but never completely hidden by non-target images. Crucially, this also guarantees that exemplars are never covered by non-target objects which could alter the desired target cluster.
While the resulting Self-Collages exhibit pasting artefacts (see \Cref{fig:self_collage_samples}), previous work has shown that eliminating these artefacts by applying a blending method does not improve the performance on downstream tasks such as object detection~\citep{dvornik2018modeling} and instance segmentation~\citep{ghiasi2021simple,zhao2022x}. Hence, we focus on simple pasting to keep the complexity of the construction process low and rely on breaking correlations between the number of artefacts and the target count of an image by pasting a constant number of images as discussed in the paper. \Cref{algo:composer} shows the pseudocode for the composer module.

If we use the ``no-overlap'' setup, which prevents objects are pasted on top of each other, and an image cannot be pasted without overlapping with an already copied image, the construction process is restarted with new random sizes and locations. To guarantee the termination of $g$, potential overlaps between images are ignored if the construction process fails 20 times.

\paragraph{Density map construction} 
We construct the density map by placing unit density at the centre of each pasted target image $\textit{\textbf{I}}\in \mathcal{I}_0$. Following \citet{djukic2022low}, we apply a Gaussian filter to blur the resulting map. Its kernel size and standard deviation vary per image and are based on the average bounding box size divided by 8.

\begin{algorithm}[H]
    \caption{The composer module}
    \label{algo:composer}
    {\scriptsize
    \begin{algorithmic}
    \State \textbf{Require:}\ \ \ $t_{\text{min}}$\Comment{minimum number of clusters}
                \State \algorithmindent $t_{\text{max}}$\Comment{maximum number of clusters}
                \State \algorithmindent $n_{\text{min}}$\Comment{minimum number of target objects}
                \State \algorithmindent $n_{\text{max}}$\Comment{maximum number of objects}
                \State \algorithmindent $K$ \Comment{total number of clusters}
                \State \algorithmindent $E$ \Comment{number of exemplars}
                \State \algorithmindent $\mathcal{O}$\Comment{set of object images}
                \State \algorithmindent $\mathcal{B}$\Comment{set of background images}
        \Statex
        \hrulefill %
        \Statex \textbf{\# create clusters and select cluster sizes}
        \State $\mathcal{C}\gets \text{k-means}_K{[(d(\mathcal{O}))^{\texttt{CLS}}]}$ \Comment{cluster $\mathcal{O}$ using $K$ clusters}
        \State $n_c \sim U[t_{\text{min}}, t_{\text{max}}]$ \Comment{ select the number of clusters}
        \State $n_0=n\sim U[n_{\text{min}}, n_{\text{max}}-n_c+1]$ \Comment{select the number of objects in the target cluster}
        \For{$i$ in range(1, $n_c-1$)} \Comment{select the number of objects in the other clusters}
            \State $n_i\sim U\left[1,n_{\text{max}}-\overbrace{\sum_{j=0}^{i-1}n_j}^{\text{previous clusters}}-\underbrace{n_c+i+1}_{\text{remaining clusters}}\right]$
       \EndFor
       \State $n_{n_c-1}\gets n_{\text{max}}-\sum_{i=0}^{n_c-2}n_i$ \Comment{set the number of objects in the final cluster}
       \Statex 
       \hrulefill %
       \Statex \textbf{\# select images}
       \For{$i$ in range(0, $n_c$)}
            \State $c_i\gets $ random\_cluster($\mathcal{C}\setminus \bigcup_{j=0}^{i-1} \{c_j\}$)  \Comment{select a random, unique cluster as the $i^{\text{th}}$cluster}
            \State $\mathcal{I}_i \gets$ select($n_i$,$c_i$,$\mathcal{O}$)  \Comment{select $n_i$ random images in cluster $c_i$ from $\mathcal{O}$}
       \EndFor
       \State $\mathcal{I} \gets \bigcup_{i=0}^{n_c-1} \mathcal{I}_i$
       \State $\boldsymbol{\mathit{\tilde I}} \gets$ select(1, $\mathcal{B}$) \Comment{select a random background image from $\mathcal{B}$}
       \Statex 
       \hrulefill %
       \Statex  \textbf{\# compose the image and pseudo ground-truth}
       \State $\textbf{B}\gets$ [\ ] \Comment{initialise an empty list for the object bounding boxes}
       \For{$\textit{\textbf{I}}$ in $\mathcal{I}$} \Comment{iterate over all object images}
            \State $s\gets$ get\_random\_size($\textit{\textbf{I}}$, $\mathcal{I}$) \Comment{get a random size, which is correlated for all images in $\mathcal{I}$}
            \State $p\gets$ get\_random\_position($s$) \Comment{get a random position for the current image}
            \State $\textit{\textbf{I}}_{r}\gets$ resize($\textit{\textbf{I}}$, $s$) \Comment{resize $\textit{\textbf{I}}$, if using segmentations, this involves cutting the object}
            \State $\boldsymbol{\mathit{\tilde I}}\gets$ paste($\boldsymbol{\mathit{\tilde I}}$, $\textit{\textbf{I}}_\text{r}$, $p$) \Comment{paste $\textit{\textbf{I}}_\text{r}$ into $\boldsymbol{\mathit{\tilde I}}$ at position $p$, if $\textit{\textbf{I}}$ is a target object, place it on top}
            \If {$\textit{\textbf{I}} \in \mathcal{I}_0$}
                \State $\textbf{b}\gets$box($p$, $s$) \Comment{create the bounding box of the current object if $\textit{\textbf{I}}$ is a target object}
                \State $\textbf{B}$.append($\textbf{b}$) \Comment{append the bounding box to the list of all object boxes}
            \EndIf
       \EndFor
       \State $\mathcal{S} \gets$ crop\_exemplars($\boldsymbol{\mathit{\tilde I}}$, $\textbf{B}$, $E$) \Comment{create $E$ exemplar crops}
       \State $\textbf{y}\gets$ create\_density\_map($\textbf{B}$) \Comment{create the density map}
       \State \textbf{return} $\boldsymbol{\mathit{\tilde I}}$, $\mathcal{S}$, $\textbf{y}$
    \end{algorithmic}
    }
\end{algorithm}

\subsection{Connected components baseline}\label{sec:connected_components_appendix}
We evaluate the connected components baseline on the FSC-147 training set to find the best values for its three thresholds $p_\text{att}$, $n_\text{head}$, and $p_\text{size}$. To this end, we run an exhaustive grid search by testing all combinations of the following settings:
\begin{align}
    p_\text{att} &\in \{0.1, 0.2, ..., 0.9, 0.95, 0.99\},\\
    n_\text{head} &\in \{1, ..., 12\},\\
    p_\text{size} &\in \{0, 0.01, 0.02, 0.05, 0.1, 0.2\}
\end{align}
This results in 792 configurations, where we pick the setup with the lowest MAE on the whole FSC-147 training set which results in the following thresholds: $p_\text{att}=0.7$, $n_\text{head}=10$, and $p_\text{size}=0$.

\subsection{Inference details}\label{sec:inference_details}
We follow \citet{liu2022countr} in our evaluation procedure.
Specifically, we resize the inference image to a height of $384$ keeping the aspect ratio fixed and scan the resulting image with a window of size $384\times 384$ and a stride of 128 pixels. The density maps of the overlapping regions are averaged when aggregating the count of the entire image.
If the image contains very small objects, defined as at least one exemplar with a width and height of less than 10 pixels, the image is divided into a $3\times3$ grid. Each of the 9 tiles is then resized to a height of $384$ pixels and processed independently. The prediction for the original image is obtained by combining the individual predictions. 
Additionally, we apply the same test-time normalisation: We normalise the predicted count by the average sum of the density map areas that correspond to the exemplars if it exceeds a threshold of 1.8.

\subsection{Self-supervised semantic counting details}\label{sec:self_supervised_counting_details}

\begin{figure}[ht]
    \centering
    \includegraphics[width=\textwidth]{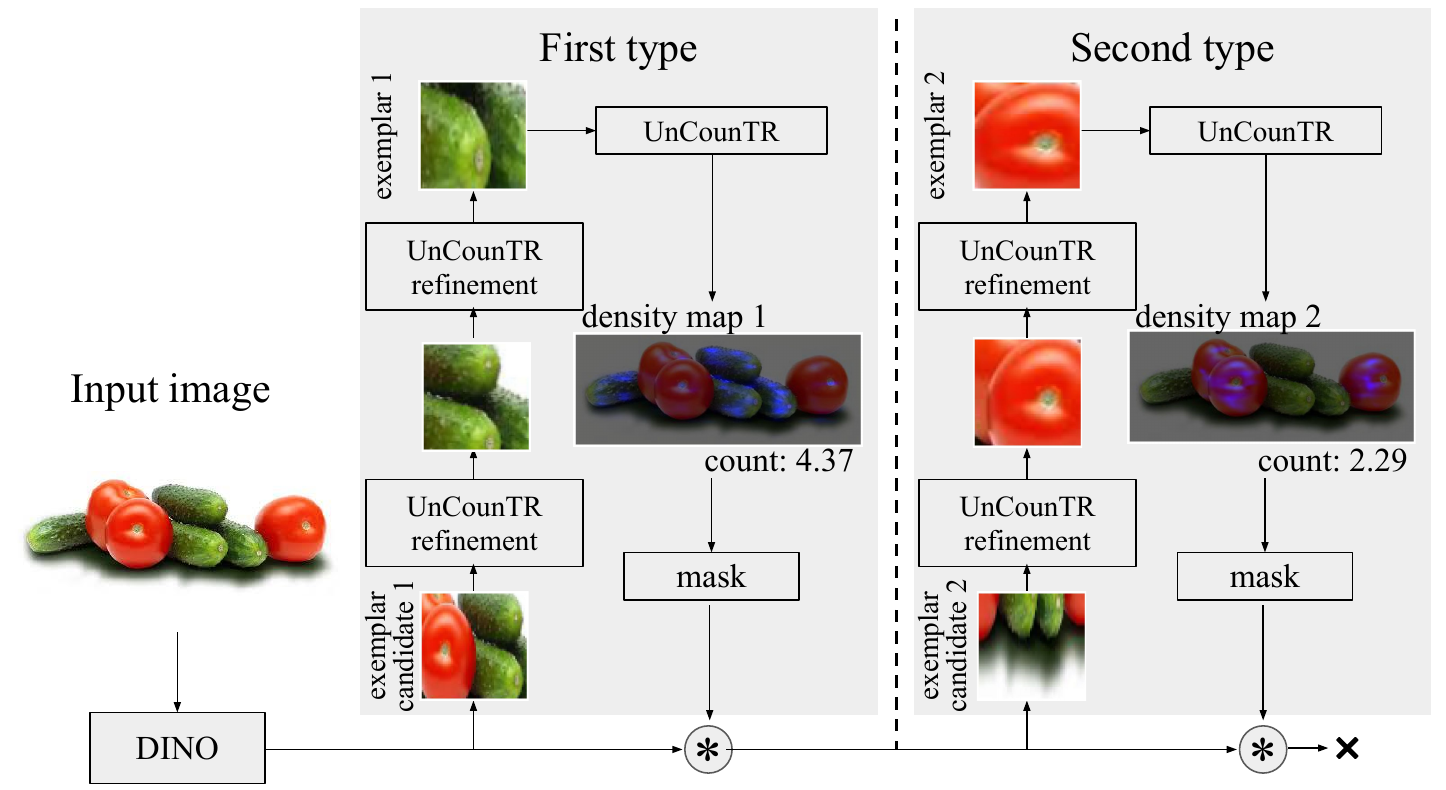}
    \caption{\textbf{Self-supervised semantic counting.} To predict the number of objects without any prior, the model uses its DINO backbone to get initial exemplar candidates, which it subsequently refines and uses to predict density maps for the discovered object types.}
    \label{fig:self_exemplar_detailed}
\end{figure}

\paragraph{Problem overview}
In the self-supervised semantic counting setup, the model automatically identifies the number of object types and appropriate exemplars to predict the density map for each category. \Cref{fig:self_exemplar_detailed} illustrates this process.

\paragraph{Processing the input image}
We first obtain the feature map of the input image $\textit{\textbf{I}}$ resized to $384\times384$ pixels using the image encoder: $\textbf{x}=\Phi(\textit{\textbf{I}})\in \mathbb{R}^{h \times w \times d}$. In addition, we extract the final \texttt{CLS}-attention map $\textbf{a}\in \mathbb{R}^{h\times w}$ where we take the average across the attention heads and only consider the self-attention for the $h\cdot w$ patches ignoring the \texttt{CLS}-token itself.

\paragraph{Proposing an exemplar candidate}
After processing the input image, the model determines an exemplar candidate to predict the density map $\mathbf{\hat{y}}^{(t)}$ of the first type $t=1$. To achieve this, we blur $\textbf{a}$ by applying a Gaussian kernel of size 3 with sigma 1.5 resulting in $\textbf{a}^{(1)}$ and identify the position $(y_\text{max}^{(1)}, x_\text{max}^{(1)}) = \argmax_{(y,x)} \textbf{a}_{y,x}^{(1)}$ of the patch with the maximum attention. Subsequently, we compute a binary feature map $\textbf{b}^{(1)} = \textbf{a}^{(1)} > 0.5 \cdot \max \textbf{a}^{(1)}$ and obtain the connected component which includes the position $(y_\text{max}^{(1)}, x_\text{max}^{(1)})$. The crop of $\textit{\textbf{I}}$ corresponding to this component is taken as the exemplar candidate $\textit{\textbf{E}}_\text{candidate}^{(1)}$. 

\paragraph{Refining the exemplar}
Using $\textit{\textbf{I}}$ and $\textit{\textbf{E}}_\text{candidate}^{(1)}$, we predict a density map
\begin{equation}\label{eq:prediction}
    \mathbf{\hat{y}}_{\text{candidate}}^{(1)}=f_\Theta(\textit{\textbf{I}}, \{\textit{\textbf{E}}_\text{candidate}^{(1)}\}),
\end{equation}
where we can skip the image encoder $\Phi$ by reusing the feature map $\textbf{x}$. $\mathbf{\hat{y}}_{\text{candidate}}^{(1)}$ is used to obtain a refined exemplar $\textit{\textbf{E}}_\text{refined}^{(1)}$. To this end, we binarise the density map $\mathbf{\hat{y}}_{\text{candidate}}^{(1)}$ by selecting values greater than 20\% of its maximum and receive the second largest connected component, where we assume the largest component represents the background and the second largest corresponds to an object of the first type. We take a crop $\textit{\textbf{E}}_\text{crop}^{(1)}$ of $\textit{\textbf{I}}$ at the location of this component. Since $\textit{\textbf{E}}_\text{crop}^{(1)}$ might contain adjacent objects of different categories, we construct $\textit{\textbf{E}}_\text{refined}^{(1)}$ by taking another center crop of $\textit{\textbf{E}}_\text{crop}^{(1)}$ reducing each dimension by a factor of 0.3. The refinement step can be repeated multiple times by using $\textit{\textbf{E}}_\text{refined}^{(1)}$ as the new candidate in \Cref{eq:prediction}. In practice, we use two refinement iterations.

\paragraph{Predicting the count}
We employ test-time normalisation and take the density map $\mathbf{\hat{y}}^{(1)}$ obtained using \Cref{eq:prediction} and replacing $\textit{\textbf{E}}_\text{candidate}^{(1)}$ with $\textit{\textbf{E}}_\text{refined}^{(1)}$ as final prediction for the first type.

\paragraph{Counting multiple categories}
To obtain predictions for more categories, we repeat these steps starting with a new maximum $(y_\text{max}^{(2)}, x_\text{max}^{(2)})$ after masking out the patches in $\textbf{a}^{(1)}$ which correspond to the current category resulting in a new attention map $\textbf{a}^{(2)}$. We identify these patches using two heuristics: First, we mask out patches where $\mathbf{\hat{y}}_{\text{candidate}}^{(1)}$ or $\mathbf{\hat{y}}_{\text{refined}}^{(1)}$, resized to match the dimensions of $\textbf{a}^{(1)}$, predict a value higher than or equal to 0.5. Second, we set the attention to 0 in an area of $5\times5$ patches centred around $(y_\text{max}^{(1)}, x_\text{max}^{(1)})$ to prevent $(y_\text{max}^{(2)}, x_\text{max}^{(2)})=(y_\text{max}^{(1)}, x_\text{max}^{(1)})$. Since every object should only be counted once and to facilitate the knowledge of previous iterations, we subtract the sum of the density maps of previous iterations, from the current prediction:
\begin{equation}
    \mathbf{\hat{y}}^{(t)} = \max\left(f_\Theta(\textit{\textbf{I}}, \{\textit{\textbf{E}}_\text{candidate}^{(t)}\}) - \sum_{t'=1}^{t-1}\max\left(\mathbf{\hat{y}}^{(t')}, \textbf{0}\right), \textbf{0}\right)
\end{equation}
This procedure keeps detecting exemplars and making predictions for new categories $t$ until the maximum remaining attention value is less than 20\% of the original maximum at which point we assume that all salient object types have been detected.

\paragraph{Evaluating the importance of the refinement steps}
Especially the first refinement step is crucial to obtain meaningful exemplars as illustrated in \Cref{fig:self_exemplar_detailed}. While the initial candidates, which are only based on the DINO backbone, successfully highlight the salient objects, they fail to focus on a single object type. A single refinement step based on UnCounTR solves this issue which indicates the importance of UnCounTR's self-supervised training for this task. The second refinement step has a more subtle impact on the exemplar quality by reducing the number of objects in each exemplar. Based on these self-supervised exemplars, UnCounTR produces counts close to the true number of objects.

\paragraph{Limitation}
While these results are promising, they are only a qualitative exploration and are intended to highlight a potential avenue for future work. 
The creation of an evaluation dataset for the semantic counting task as well as the development of a metric to measure exemplar quality are required for a more thorough evaluation of this use case.

\section{More details of Self-Collages}\label{sec:self_collage_details}

\subsection{Underlying assumptions for Self-Collages}\label{sec:assumptions}

In the paper, we describe the construction of Self-Collages with ImageNet-1k and SUN397 dataset. Our underlying assumptions are twofold:
\begin{enumerate}
    \item Images in the SUN397 dataset do not contain objects to serve as the background for our Self-Collages.
    \item Images in the ImageNet-1k dataset feature a single salient object to obtain correct pseudo labels.
\end{enumerate}

\begin{figure}[ht]
    \centering
    \includegraphics[width=\textwidth]{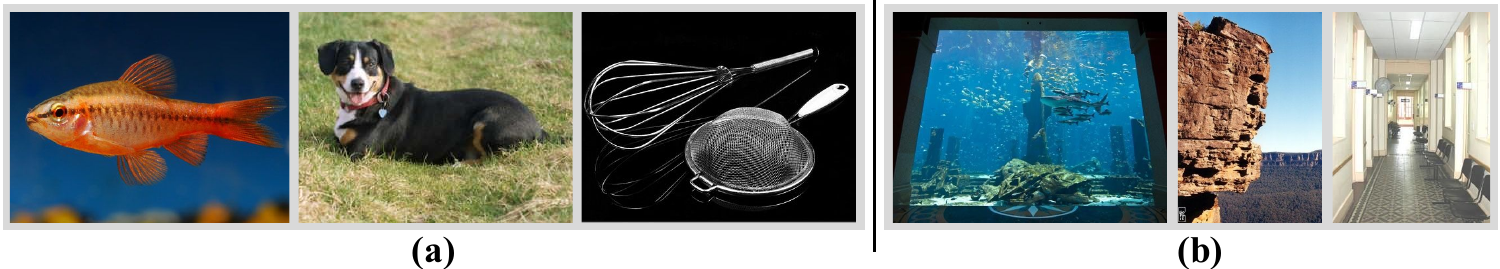}
    \caption{\textbf{ImageNet-1k and SUN397 images.} \textbf{(a)} Example images from the ImageNet-1k dataset~\citep{russakovsky2015imagenet}.
    \textbf{(b)} Example images from the SUN397 dataset~\citep{xiao2010sun}.
    While the figures in the SUN397 dataset may contain multiple objects, there is no clearly salient object.}
    \label{fig:imagenet_sun_samples}
\end{figure}

\Cref{fig:imagenet_sun_samples} shows three samples from ImageNet-1k and SUN397. We can see that even though images in the latter do contain objects, see \eg the fish in the aquarium, they are usually not salient. Hence, the first assumption is still reasonable for constructing Self-Collages. We acknowledge that this assumption has its limitations and introduces noises for Self-Collages,~\eg salient objects exist in some images from SUN397. In \Cref{sec:dataset_ablations}, we consider variants of the default setup to investigate the robustness of our method against violations of this assumption.

The second assumption is crucial to derive a strong supervision signal from unlabelled images. While some ImageNet-1k images contain multiple salient objects, see \eg the right-most image in Subfigure~\textbf{a}, the final performance of our method shows that the model is able to learn the task even with this noisy supervision. Some techniques such as filtering images based on their segmentation masks, might be able to further improve the supervision signal. We consider these methods as future works.

\subsection{Examples of Self-Collages}\label{sec:self_collage_examples}

\begin{figure}[ht]
    \centering
    \includegraphics[width=0.9\textwidth]{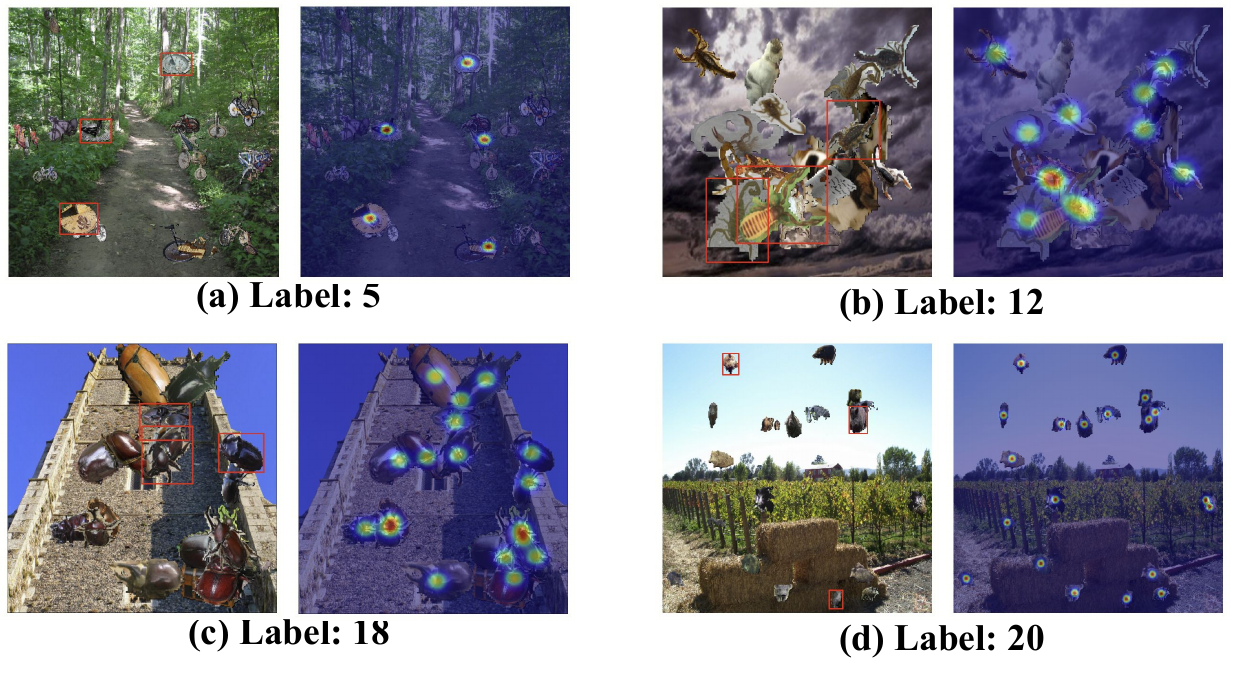}
    \caption{\textbf{Example Self-Collages.} The Subfigures show Self-Collages for different counts. The red boxes indicate the exemplars and the heatmaps show the pseudo ground-truth density maps. }
    \label{fig:self_collage_samples}
\end{figure}

\Cref{fig:self_collage_samples} shows different examples of Self-Collages. While the whole pipeline does not rely on any human supervision, the Self-Collages contain diverse types that align with human concepts from objects like bicyles (Subfigure~\textbf{a}) to animals such as beetles (Subfigure~\textbf{c}). Likewise, the unsupervised segmentation method proposed by \citet{shin2022unsupervised} successfully creates masks for the different instances. The correlated sizes lead to similarly sized objects in each Self-Collage with some samples containing primarily small instances (Subfigures~\textbf{a} and \textbf{d}) and others having mainly bigger objects (Subfigures~\textbf{b} and \textbf{c}). Adding to the diversity of the constructed samples, some Self-Collages show significantly overlapping objects (Subfigure~\textbf{b}) while others have clearly separated entities (Subfigure~\textbf{d}). Since the total number of pasted objects is constant, the amount of pasting artefacts in each image does not provide any information about the target count. The density maps indicate the position and number of target objects, overlapping instances result in peaks of higher magnitude (Subfigure~\textbf{b} and \textbf{c}). By computing the parameters of the Gaussian filter used to construct the density maps based on the average bounding box size, the area covered by density mass correlates with the object size in the image.

\subsection{Collages in other works}

The idea of deriving a supervision signal from unlabelled images by artificially adding objects to background images, the underlying idea behind Self-Collages, is also used in other domains such as object discovery and segmentation.

\citet{arandjelovic2019object} propose a generative adversarial network, called copy-pasting GAN, to solve this task in an unsupervised manner. They train a generator to predict object segmentations by using these masks to copy objects into background images. The learning signal is obtained by jointly training a discriminator to differentiate between fake, \ie images containing copied objects, and real images. A generator that produces better masks results in more realistic fake images and has therefore higher chances of fooling the discriminator. During inference, the generator can be used to predict instance masks in images. By contrast, our composer module $g$ is only used during training to construct samples. Hence, we do not need an adversarial setup and, without the need to update $g$ during training, the composer module does not have to be differentiable.

To boost the performance of instance segmentation methods, \citet{zhao2022x} propose the X-Paste framework. It leverages recent multi-modal models to either generate images for different object classes from scratch or filter web-crawled images. After creating instance masks and filtering these objects, the resulting images are pasted into background images. Generated samples can then be used in isolation or combined with annotated samples to train instance segmentation methods. Compared to this work, our goal is not to train instance segmentation but counting methods where the correct number of objects in the training samples is more important than the quality of the individual segmentations. Partly due to this, our method is conceptually simpler than X-Paste and does not require, for example, multiple filtering steps.

\section{Further results\label{sec:further_results}}

In this section, we further analyse UnCounTR's performance in out-of-domain settings (see \Cref{sec:generatlisation_ood}) and explore ways to improve UnCounTR's default setup based on recent advances in self-supervised representation learning (see \Cref{sec:improvements}). 
We then evaluate our method under different data distributions in \Cref{sec:dataset_ablations}.
We conclude this section with a qualitative comparison between UnCounTR and FasterRCNN (see \Cref{sec:comparison_fasterrcnn_dino_countr}) and show additional qualitative results (see \Cref{sec:qualitative_results}).

\subsection{Generalisation to out-of-domain count distributions\label{sec:generatlisation_ood}}

In this section, we examine UnCounTR's performance on the different FSC-147 subsets, as presented in the paper, to investigate its generalisation capabilities to new count distributions. We consider the results on the three subsets \textit{low}, \textit{medium}, and \textit{high} with an average of 12, 27, and 117 objects per image respectively. Since UnCounTR is trained with Self-Collages of 11 target objects on average, the count distribution of FSC-147 \textit{low} can be seen as in-distribution while the \textit{medium} and \textit{high} subsets are increasingly more out-of-distribution (OOD).

Unsurprisingly, the model performs worse on subsets whose count distributions deviate more from the training set. Looking at FSC-147 \textit{medium}, the RMSE changes only slightly considering the significant increase in the number of objects compared to \textit{low}. When moving to the \textit{high} subset whose count distribution differs significantly from the training set, the error increases substantially.

\begin{figure}[ht]
    \centering
    \begin{subfigure}{0.49\textwidth}
        \centering
        \includegraphics[width=\textwidth]{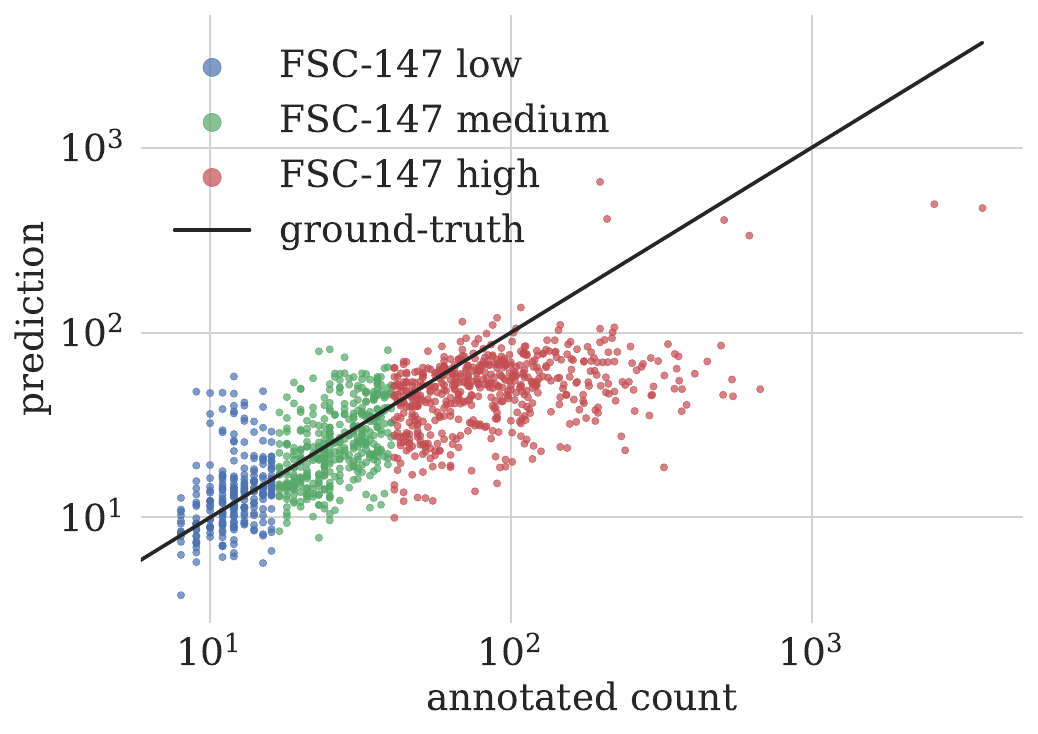}
        \caption{}
        \label{fig:dino_countr_abs_predictions}
    \end{subfigure}%
    \hfill
    \begin{subfigure}{0.49\textwidth}
        \centering
        \includegraphics[width=\textwidth]{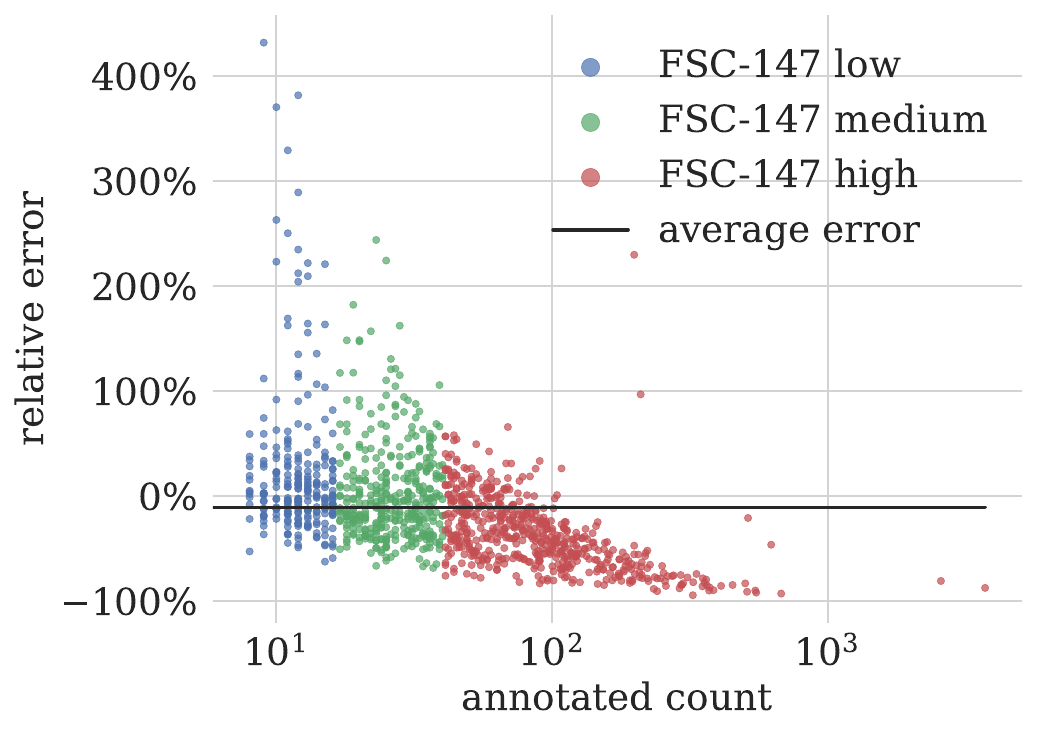}
        \caption{}
        \label{fig:dino_countr_rel_error}
    \end{subfigure}
    \caption{\textbf{UnCounTR's predictions.} \Cref{fig:dino_countr_abs_predictions} compares the model's predictions on the different FSC-147 test subsets with the ground-truth. \Cref{fig:dino_countr_rel_error} visualises the relative error for these predictions.}
    \label{fig:dino_countr_predictions}
\end{figure}
These trends can also be seen in \Cref{fig:dino_countr_predictions}. \Cref{fig:dino_countr_abs_predictions} shows that the model's predictions are distributed around the ground-truth for FSC-147 \textit{low} and \textit{medium}. On the \textit{high} subset, the model tends to underestimate the number of objects in the images. The relative error becomes increasingly negative as these counts increase (see \Cref{fig:dino_countr_rel_error}) illustrating the limitations of generalising to OOD count ranges.
\begin{table*}[tb]
    \centering
    \small
    \label{tabl:FSC147_subsets}
    \setlength{\tabcolsep}{0.4em}
    \begin{tabular}{l @{\hskip 0.5em} ccc c ccc c ccc }
        \toprule
        \multirow{2}*{Method}  & \multicolumn{3}{c}{FSC-147 \textit{low}}  && \multicolumn{3}{c}{FSC-147 \textit{medium}} && \multicolumn{3}{c}{FSC-147 \textit{high}} \\
        \cmidrule{2-4} \cmidrule{6-8} \cmidrule{10-12} 
        & MAE$\downarrow$ & RMSE$\downarrow$ & $\tau\uparrow$ && MAE$\downarrow$ & RMSE$\downarrow$ & $\tau\uparrow$ &&  MAE$\downarrow$ & RMSE$\downarrow$ & $\tau\uparrow$\\
        \midrule
        \textbf{UnCounTR (ours)}  & {5.60} & 10.13 & {0.27} && {9.48} & {12.73} & {0.34} && {67.17} & {189.76}  & {0.26}\\
        \quad $\sigma$(5 runs)  &\,${\pm0.48}$ & $\,{\pm0.84}$ & $\,{\pm0.02}$ && $\,{\pm0.19}$ & $\,{\pm0.33}$ & $\,{\pm0.02}$ && $\,{\pm1.03}$ & $\,{\pm1.38}$  & $\,{\pm0.01}$\\
        \bottomrule
    \end{tabular}
    \caption{\textbf{UnCounTR's performance on FSC-147.} Evaluation on different FSC-147 test subsets as described in the paper.}
    \vspace{-1mm}
\end{table*}

While this is expected, we can assume that a model that learned a robust notion of numerosity gives higher count estimates for images containing more objects even in these OOD settings. This can be quantified using the rank correlation coefficient Kendall's $\tau$. Interestingly, the correlation coefficient on FSC-147 \textit{low} and \textit{high} is almost the same, the highest value for $\tau$ is achieved on the \textit{medium} subset (see \Cref{tabl:FSC147_subsets}). This suggests that while UnCounTR's count predictions become less accurate for out-of-distribution samples, the model's concept of numerosity as learned from Self-Collages generalises well to much higher count ranges.

\subsection{Improving upon UnCounTR\label{sec:improvements}}

\begin{table}[tb]
    \centering
    \begin{tabular}{ l c c @{\hskip 0.5em} ccc c ccc }
        \toprule
          &  &  & \multicolumn{3}{c}{FSC-147}  && \multicolumn{3}{c}{FSC-147 \textit{low}}\\
        \cmidrule{4-6} \cmidrule{8-10}
        Backbone & similarity & refinement & MAE$\downarrow$ & RMSE$\downarrow$ & $\tau\uparrow$ && MAE$\downarrow$ & RMSE$\downarrow$ & $\tau\uparrow$\\
        \hline
        DINOv2 & \xmark & \xmark & 34.35 & 132.02 & 0.63 && 4.10 & 7.34 & 0.32\\
        DINOv2 & \cmark & \xmark & 31.06 & 119.35 & 0.67 && 3.98 & 7.58 & \textbf{0.33}\\
        DINOv2 & \cmark & \cmark & \textbf{28.67} & \textbf{118.40} & \textbf{0.71} && \textbf{3.88} & \textbf{7.03} & \textbf{0.33} \\
        \rowcolor{lightgray} DINO & \xmark & \xmark &  {35.77} & 130.34 & 0.57  && {5.60} & {10.13} & {0.27}\\
        \bottomrule
    \end{tabular}
    \caption{\textbf{Improving UnCounTR's performance.} We explore several variations of UnCounTR's default setup, highlighted in \colorbox{lightgray}{grey}, to further close the performance gap to its supervised counterparts.}
    \label{tabl:FSC147_improvements}
\end{table}

In the paper, we introduce UnCounTRv2 which integrates three modifications compared to the default setup of UnCounTR based on the very recent DINOv2~\citep{oquab2023dinov2} backbone. Building on this change, we investigate two further modifications: exploiting cluster similarity and refining the model's predictions. \Cref{tabl:FSC147_improvements} shows the results for the different variations.

\subsubsection{Updating the backbone}

First, we update the DINO backbone~\citep{caron2021emerging} with the newer DINOv2~\citep{oquab2023dinov2}. More specifically, we employ the ViT-B model with a patch size of 14. Due to the different patch size, we change the resolution of the exemplars slightly from $64\times64$ to $70\times70$. In addition, we update the sliding window's dimension to $392\times 392$ during inference.

The results in \Cref{tabl:FSC147_improvements} show that updating the backbone improves the overall performance on the whole FSC-147 dataset, with a small increase in RMSE by 1.3\% being the only exception. In particular, this change seems to be beneficial for images with lower counts, the corresponding metrics improve by 19-28\%. This demonstrates UnCounTR's ability to take advantage of recent advances in self-supervised representation learning as discussed in the paper.

\subsubsection{Cluster similarity}

\begin{figure}[ht]
    \centering
    \includegraphics[width=0.9\textwidth]{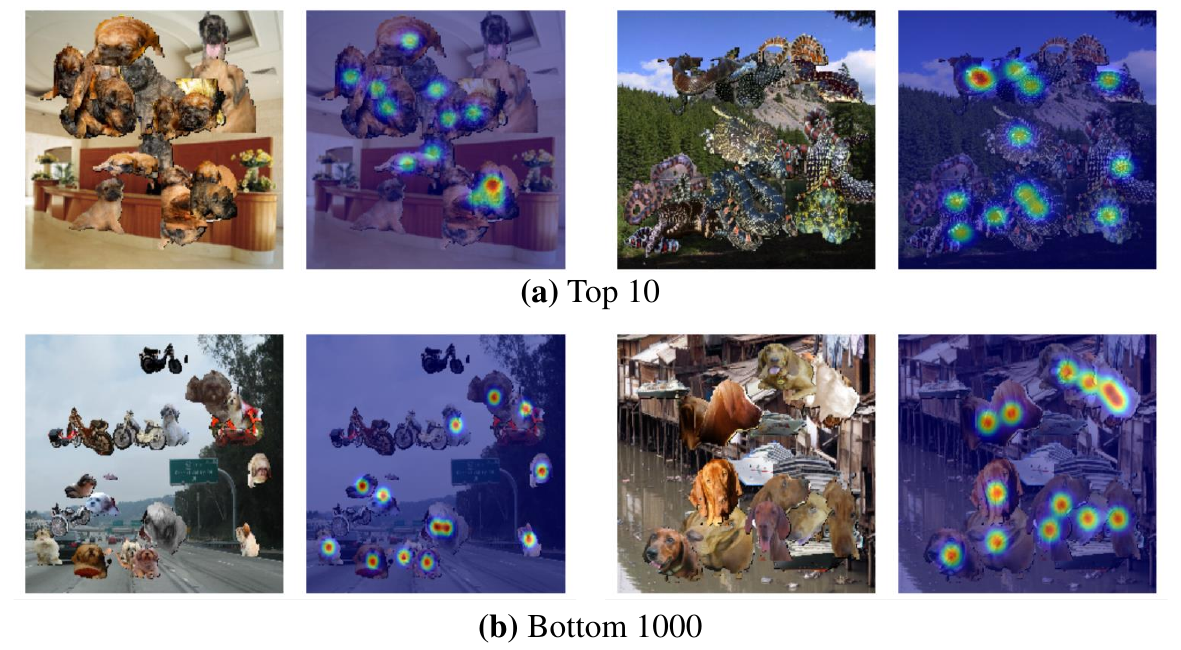}
    \caption{\textbf{Constructing Self-Collages based on cluster similarity.} The similarity between clusters can be used when constructing Self-Collages, the pseudo ground-truth on the right of each image indicates the objects of the target cluster. Subfigure~\textbf{a} shows Self-Collages where the non-target cluster was randomly chosen among the 10 clusters most similar to the target cluster. Due to the high similarity, objects in the target and non-target clusters are not easily distinguishable. By contrast, the examples in Subfigure~\textbf{b} use the 1000 most different clusters. The two types in each image are visually very distinct.}
    \label{fig:cluster_sim}
\end{figure}

Since multiple object types are pasted to generate our training images, we next exploit information about their similarity. We hypothesise that increasing the difficulty of the counting task during training by selecting non-target clusters that are similar to the target clusters could facilitate the learning process.
This idea resembles the use of hard negatives in contrastive learning which has been shown to improve the robustness of learned representations~\citep{xuan2020hard,ge2021robust,zhang2022correct}.
Unlike in supervised setups with manually annotated classes, information about the similarity of object clusters is readily available in our self-supervised setup and can be computed simply as the negative Euclidean distance between the cluster centres. \Cref{fig:cluster_sim} shows the effect of cluster similarity on the constructed Self-Collages.

In \Cref{tabl:FSC147_similarity}, we compare different ways of exploiting the similarity information. It can be seen that selecting non-target clusters that are similar to the target cluster significantly improves the performance. Importantly, picking non-target clusters that are too similar to the target cluster severely harms the training process. Considering the qualitative samples in \Cref{fig:cluster_sim}, very similar object types are almost visually indistinguishable even for humans. Hence, we use the 100 clusters most similar to the target cluster excluding the top 10. Unlike the updated backbone, this change improves the performance especially on images with higher counts as seen in \Cref{tabl:FSC147_improvements}.

\begin{table}[tb]
    \centering
    \begin{tabular}{ c @{\hskip 0.5em} ccc }
        \toprule
          & \multicolumn{3}{c}{FSC-147} \\
        \cmidrule{2-4}
        similarity range & MAE$\downarrow$ & RMSE$\downarrow$ & $\tau\uparrow$\\
        \hline
        \xmark & 34.35 & 132.02 & 0.63\\
        top 10 & 36.67 & 130.30 & 0.59 \\
        10-100 & \textbf{31.06} & \textbf{119.35} & \textbf{0.67} \\
        bottom 1000 & 34.71 & 131.12 & 0.63 \\
        \bottomrule
    \end{tabular}
    \caption{\textbf{Using cluster similarities to construct Self-Collages.} We use cluster similarities in the composer module $g$ to select non-target clusters during Self-Collage construction. Top $X$ describes the setup where we pick the non-target cluster among the $X$ clusters that are most similar to the target cluster. Likewise, the clusters are chosen from the set of the $X$ most dissimilar clusters in the bottom $X$ setting. If we specify a range $X$-$Y$, the set of possible non-target clusters is equal to top $Y$ without the elements in top $X$. All setups use DINOv2 as backbone. }
    \label{tabl:FSC147_similarity}
\end{table}

\subsubsection{Prediction refinement}
Lastly, we modify our evaluation protocol to mitigate the count distribution shift between training and the FSC-147 test set as discussed in \Cref{sec:generatlisation_ood}. To this end, we employ a refinement strategy which aims in particular at improving the predictions for images with high object counts. First, we obtain a prediction using the default inference setup described in \Cref{sec:inference_details}. Then, if the model predicts more than 50 objects, we utilise the same setup as employed for small objects where we split the image into 9 tiles and predict the counts for each of them independently before aggregating the final prediction (see \Cref{sec:inference_details}).

\Cref{tabl:FSC147_improvements} shows that employing this evaluation protocol further improves the performance resulting in an MAE of 28.67 on FSC-147. Combining all three modifications reduces the MAE of UnCounTRv2 compared to the default UnCounTR setup by 20\% on FSC-147 and 31\% on FSC-147 \textit{low} which narrows the gap to the supervised counterparts and highlights the potential of unsupervised counting based on Self-Collages.

\subsection{Self-Collages based on different datasets} \label{sec:dataset_ablations}

In \Cref{tabl:appendix_FSC147_ablation_dataset}, we ablate the datasets used for object ($\mathcal{O}$) and background images ($\mathcal{B}$) to investigate the behaviour of our method under different data distributions. \Cref{fig:training_samples_ablations} shows training samples for the ablated setups. We first describe the different datasets in \Cref{sec:dataset_ablations_description}, followed by the ablation results in \Cref{sec:dataset_ablation_results}. 

\subsubsection{Dataset ablations}\label{sec:dataset_ablations_description}

\begin{figure}[ht]
    \centering
    \includegraphics[width=0.9\textwidth]{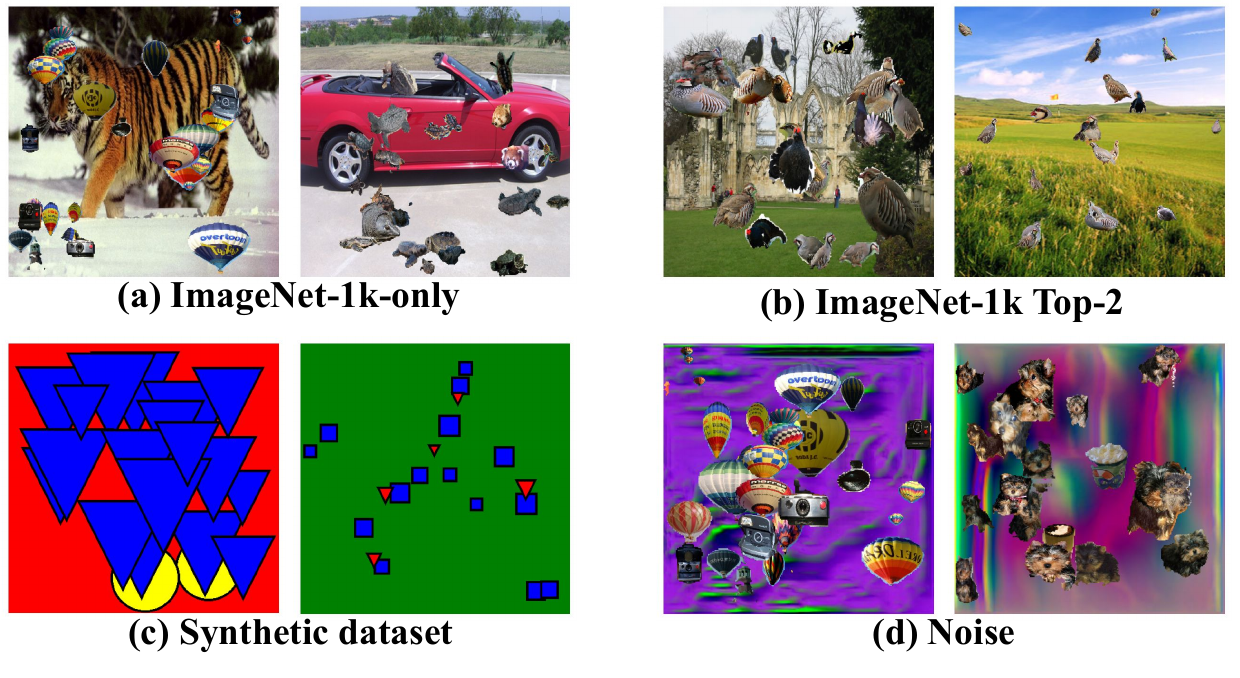}
    \caption{\textbf{Dataset ablations.} We modify the training dataset by using different datasets for $\mathcal{O}$ and $\mathcal{B}$ to investigate the effect on the final counting performance.}
    \label{fig:training_samples_ablations}
\end{figure}

\paragraph{ImageNet-1k-only} To simplify the construction process, we experiment with using the ImageNet-1k dataset for both, $\mathcal{O}$ and $\mathcal{B}$. 
To make sure that the counting task is not affected by the background, we exclude all images in the two clusters used for the object images before randomly selecting the background image.

\paragraph{ImageNet-1k Top-2} As a simpler version of the default setup, we filter the ImageNet-1k dataset to only use the images in the two biggest clusters, which we call ImageNet-1k Top-2. Looking at Subfigure~\textbf{b} in \Cref{fig:training_samples_ablations}, these clusters seem to correspond to two different types of birds. This reduces the number of images from 1,281,167 to 1,266. Because this subset only features two clusters, every Self-Collage contains the same object types.

\paragraph{Synthetic dataset} We construct a synthetic dataset based on simple shapes. To this end, we combine three types of shapes, \texttt{squares}, \texttt{circles}, and \texttt{triangles}, and four different colours, \texttt{red}, \texttt{green}, \texttt{blue}, and \texttt{yellow}, to obtain a total of 12 possible object types. After randomly picking two different types, we select a random background colour amongst the colours not used for the objects.

\paragraph{Noise} We use the StyleGAN-Oriented dataset proposed by \citet{baradad2021learning}, which is based on a randomly initialised StyleGANv2~\citep{karras2020analyzing}, as noise dataset to draw background images from. In total, it contains 1.3M synthetic images. We refer to the original work~\citep{baradad2021learning} for more details.

\subsubsection{Dataset ablation results}\label{sec:dataset_ablation_results}
\begin{table}[tb]
    \centering
    \small
    \begin{tabular}{ l l @{\hskip 0.5em} ccc c ccc }
        \toprule
          &  & \multicolumn{3}{c}{FSC-147}  && \multicolumn{3}{c}{FSC-147 \textit{low}}\\
        \cmidrule{3-5} \cmidrule{7-9}
        $\mathcal{O}$ & $\mathcal{B}$ & MAE$\downarrow$ & RMSE$\downarrow$ & $\tau\uparrow$ && MAE$\downarrow$ & RMSE$\downarrow$ & $\tau\uparrow$\\
        \hline
        \multicolumn{2}{c}{ImageNet-1k} & 35.68 & 130.10 & 0.56 && 6.78 & 12.43 & 0.25 \\
        ImageNet-1k Top-2 & SUN397 & 39.47 & 132.37 & 0.41 && 12.08 & 17.84 & 0.16 \\
        \multicolumn{2}{c}{Synthetic dataset} & 45.16 & 144.60 & 0.26 && 7.03 & 11.04 & 0.23 \\
        ImageNet-1k & Noise & \textbf{34.43} & \textbf{128.73} & \textbf{0.60} && 5.94 & 10.72 & 0.26 \\
        \rowcolor{lightgray} ImageNet-1k & SUN397 &  {35.77} & 130.34 & 0.57  && \textbf{5.60} & \textbf{10.13} & \textbf{0.27}\\
        \bottomrule
    \end{tabular}
    \caption{\textbf{Using different datasets to construct Self-Collages.} We vary the datasets used by the composer module $g$ to obtain object ($\mathcal{O}$) and background images ($\mathcal{B}$) when constructing Self-Collages. The default setting is highlighted in \colorbox{lightgray}{grey}.}
    \label{tabl:appendix_FSC147_ablation_dataset}
\end{table}

\paragraph{Image diversity improves generalisability}  It can be seen, that gradually decreasing the image diversity, by using the same dataset for $\mathcal{O}$ and $\mathcal{B}$, using only a very small ImageNet-1k subset for $\mathcal{O}$, or using a fully synthetic dataset, harms the performance on the FSC-147 dataset.

\paragraph{Simple objects are sufficient for low counts only} While using the fully synthetic dataset yields the worst results on FSC-147, the performance on the FSC-147 \textit{low} subset is comparable to the setup using only ImageNet-1k. This indicates the importance of more realistic objects for the generalisability to higher counts. At the same time, synthetic objects seem to be sufficient for learning to predict the number of objects in images with few instances.

\paragraph{Synthetic but diverse backgrounds perform similarly to real images} When replacing SUN397 with a noise dataset, the performance on the whole FSC-147 dataset improves while being slightly worse on the FSC-147 \textit{low} subset.

\paragraph{Self-Collages are robust against violations of the background assumption} Comparing the setup which uses the ImageNet-1k dataset for $\mathcal{O}$ and $\mathcal{B}$ to the default setting, we can see that even by explicitly violating the assumption that there are no salient objects in $\mathcal{B}$, the performance is not significantly affected. However, by using a noise dataset as $\mathcal{B}$ where the assumption holds, we can further improve the performance on FSC-147. This indicates that while our method is robust to violations of the aforementioned assumption, using artificial datasets where the assumption is true, can be beneficial. 

\subsection{Comparing FasterRCNN and UnCounTR}\label{sec:comparison_fasterrcnn_dino_countr}
\begin{figure}[H]
\centering
\includegraphics[width=1.0\textwidth]{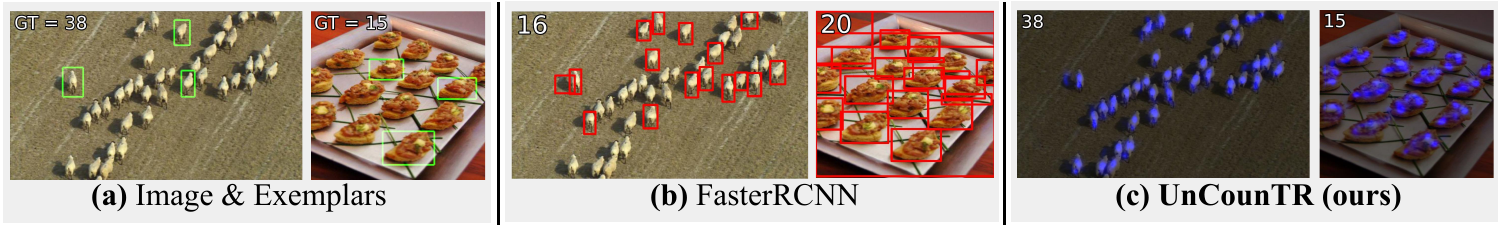}
    \caption{\textbf{When UnCounTR works better than FasterRCNN.} Given two images with exemplars \textbf{(a)} we compare the output predictions of FasterRCNN \textbf{(b)} and of our model \textbf{(c)}. 
    We find that FasterRCNN either misses most objects in high-density settings or detects non-target instances which is because the model cannot utilise any prior knowledge in the form of exemplars.
    \label{fig:rcnn_comparison}}
    \vspace{-3mm}
\end{figure}

\subsection{Qualitative results}\label{sec:qualitative_results}

\begin{figure}[h!]
    \centering
    \includegraphics[width=\textwidth]{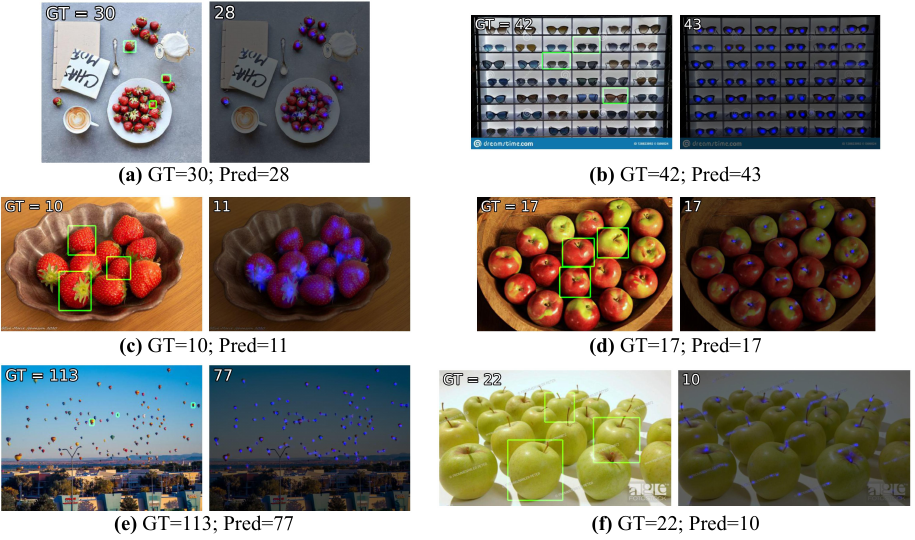}
    \caption{\textbf{Qualitative UnCounTR results.} We show UnCounTR's predictions on six different images from the FSC-147 test set. The green boxes represent the exemplars and the number in the top-left of each image indicate the ground-truth and predicted count for each sample. Our prediction is the sum of the heatmap rounded to the nearest integer.}
    \label{fig:appendix_qualitative}
\end{figure}

\Cref{fig:appendix_qualitative} visualises predictions of our model UnCounTR on the FSC-147 dataset. The model predicts good count estimates even for images with more than twice as many objects as the maximum number of objects seen during training (Subfigure~\textbf{b}). In general, the model successfully identifies the object type of interest and focuses on the corresponding instances even if they only make up a small part of the entire image (Subfigure~\textbf{a}). However, UnCounTR still misses some instances in these settings.

For very high counts (Subfigure~\textbf{e}) and images with artefacts such as watermarks (Subfigure~\textbf{f}), the model sometimes fails to predict a count close to the ground-truth.

\end{document}